\documentclass[journal, dvipsnames]{IEEEtran}

\usepackage{amsmath}
\usepackage{amssymb}
\usepackage{bm}
\usepackage{siunitx}
\usepackage[pdftex]{graphicx}
\usepackage{enumerate}

\usepackage{xspace}
\usepackage[super]{nth}

\usepackage{cite}
\usepackage{url}
\usepackage[hidelinks]{hyperref}
\usepackage{orcidlink}

\DeclareMathOperator*{\argmin}{argmin}

\DeclareSIUnit\costs{\text{cost-units}}

\newcommand{\sota}{state-of-the-art\xspace}

\newcommand{\resvec}[2]{{_\mathcal{\scriptscriptstyle #2} \bm{#1}}}
\newcommand{\resveccomp}[3]{{_\mathcal{\scriptscriptstyle #2} #1 _{#3}}}

\newcommand{\floor}[1]{\left\lfloor #1 \right\rfloor}

\begin{document}
\title{High-Speed, All-Terrain Autonomy:\\Ensuring Safety at the Limits of Mobility}

\author{
    James~R.~Baxter\,\orcidlink{0009-0004-7232-5645}, Bogdan~I.~Epureanu\,\orcidlink{0000-0002-1710-9278}, Paramsothy~Jayakumar, Tulga~Ersal\,\orcidlink{0000-0002-6811-8529}$^*$%
    \thanks{J. R. Baxter, B. I. Epureanu and T. Ersal are with the Department of Mechanical Engineering, University of Michigan, Ann Arbor, MI. (email: {jrbaxter, epureanu, tersal}@umich.edu). P. Jayakumar is with U.S. Army Ground Vehicle Systems Center, Warren, MI. (e-mail: paramsothy.jayakumar.civ@army.mil).}%
    \thanks{The authors acknowledge the technical and financial support of the Automotive Research Center (ARC) in accordance with Cooperative Agreement W56HZV-24-2-0001 U.S. Army DEVCOM Ground Vehicle Systems Center (GVSC) Warren, MI.}%
    \thanks{DISTRIBUTION STATEMENT A. Approved for public release; distribution is unlimited. OPSEC \#9732}%
    \thanks{* Corresponding author}%
}

\maketitle

\begin{abstract}
A novel local trajectory planner, capable of controlling an autonomous off-road vehicle on rugged terrain at high-speed is presented. Autonomous vehicles are currently unable to safely operate off-road at high-speed, as current approaches either fail to predict and mitigate rollovers induced by rough terrain or are not real-time feasible. To address this challenge, a novel model predictive control (MPC) formulation is developed for local trajectory planning. A new dynamics model for off-road vehicles on rough, non-planar terrain is derived and used for prediction. Extreme mobility, including tire liftoff without rollover, is safely enabled through a new energy-based constraint. The formulation is analytically shown to mitigate rollover types ignored by many state-of-the-art methods, and real-time feasibility is achieved through parallelized GPGPU computation. The planner's ability to provide safe, extreme trajectories is studied through both simulated trials and full-scale physical experiments. The results demonstrate fewer rollovers and more successes compared to a state-of-the-art baseline across several challenging scenarios that push the vehicle to its mobility limits.
\end{abstract}


\section{Introduction}
\label{sec:intro}
\IEEEPARstart{A}{ll-terrain} vehicles have tremendous mobility capacity: the physical platforms are capable of navigating rugged off-road terrain, even at high speeds.
However, for them to be useful as autonomous ground vehicles (AGVs) in applications ranging from extraterrestrial planetary exploration to mining and disaster relief,
    where vehicles are required to robustly navigate many types of terrain, including rugged off-road environments,
    the autonomous control algorithm must be advanced enough to fully utilize the mobility of the platforms.

A crucial feature of such a successful control algorithm is the ability to ensure vehicle safety.
This is especially relevant when operating in remote regions or on other planets.
Yet in the off-road domain, this intersects with the simultaneous need for vehicles to operate at high speed.
For example, a steep hill composed of loose soil can only be summited if the vehicle has sufficient speed at the base, which may be rough and uneven.
As reviewed below in Sec. \ref{sec:lit-review}, the combination of high speed, terrain roughness, and the need to ensure vehicle safety is not adequately addressed by existing methods.

When performing high-speed operations over rugged terrain, rollover poses a significant danger \cite{hanModelPredictiveControl2024},
    and it may be provoked by a variety of means:
some rollovers may be caused entirely by tire forces acting within the longitudinal-lateral plane of the vehicle,
    and others may be operate via exciting the vertical and suspension dynamics of the vehicle.
For example, the former could be caused by taking a turn too quickly, assuming there is sufficient traction,
    and the latter could be caused by suddenly driving a wheel into a berm at high speed or rapidly moving between alternating side-slopes.

\begin{figure}
    \centering
    \includegraphics[width=0.9\linewidth]{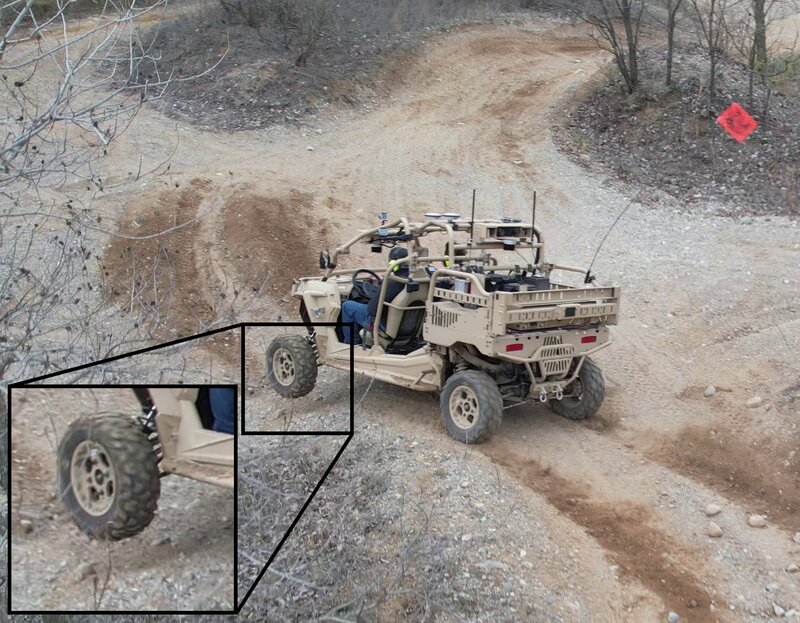}
    \caption{
        Off-road vehicles experience significant vertical dynamics (evidenced by the extreme bend of the fiberglass flagpole)
        and even tire liftoff (shown in close-up view) when traveling over the rugged terrain considered in this work, even at moderate speed (\qty{5}{\m\per\s}).
        Since these dynamics contribute to rollover, navigation, especially at high speeds, is dangerous.
        This work demonstrates a method to mitigate the risk of rollover through advanced control design.}
    \label{fig:liftoff}
\end{figure}

In rugged off-road domains, especially if an autonomous all-terrain vehicle platform is desired to operate at maximum mobility, these hazards must be addressed.
It is not sufficient to think of safety as only avoiding obstacles: these environments are characterized by an absence of flat, even ground,
    and the distinction between the nominal terrain surface and separate outcroppings is ambiguous at best (for example, the deep wheel ruts in Fig. \ref{fig:liftoff}).
A successful approach must mitigate the risk of both identified types of rollovers,
in addition to avoiding obstacles.
Planners are needed that can ensure vehicle safety while operating at the limits of mobility, and this work aims to fill that gap.
The primary original contributions are
\begin{enumerate}
    \item a detailed analysis of vehicle and constraint modeling for rollover-prevention of AGVs operating over rugged off-road terrain at high-speed,
        including identification of the requirements to ensure safety in this domain;

    \item construction of a model predictive control (MPC) formulation that meets the previously identified requirements via
        synthesis of a vehicle dynamics model with explicit consideration of 3D terrain geometry and
        construction of an energy-based rollover-prevention constraint; and

    \item closed-loop evaluation of the proposed formulation,
        including demonstration of real-time feasibility
        and comparison against a \sota baseline through both high-fidelity simulation and physical experimentation.
\end{enumerate}

Additionally, the software developed to complete this work, the AllTerrainAutonomyToolbox.jl package for Julia \cite{bezansonJuliaFreshApproach2017}, is released as free and open-source software.
    The package provides utilities for performing simulated MPC trials using the baseline and proposed formulations from this work,
    and it also includes the real-time-feasible CUDA C++ implementations,
    a high fidelity plant co-simulator,
    and a ROS1 Noetic \cite{quigleyROSOpensourceRobot} implementation of the local trajectory planner.
Source code may be found at \href{https://www.github.com/JamesRBaxter/AllTerrainAutonomyToolbox.jl}{github.com/JamesRBaxter/AllTerrainAutonomyToolbox.jl}.

The remainder of the paper is organized as follows:
\begin{itemize}
    \item Sec. \ref{sec:lit-review} conducts a literature review of related work.
    \item Sec. \ref{sec:modeling-safety}
    defines the baseline extended single-track vehicle dynamics model and
    analyzes the theory of safety-constraint formulation for autonomous vehicles on rough off-road terrain.
    \item Sec. \ref{sec:formulation}
    defines the single-level sampling-based MPC formulation proposed by this work,
    including the proposed single rigid body vehicle dynamics model.
    \item Sec. \ref{sec:setup} describes the off-road autonomous navigation tasks used to evaluate the proposed formulation.
    \item Sec. \ref{sec:results} reports and discusses results from 9900 simulated and 48 physical trials of the navigation tasks.
    \item Sec. \ref{sec:conclusion} concludes the study.
\end{itemize}

Additionally,
    App. \ref{app:notation} describes modeling notation,
    App. \ref{app:params} provides numerical parameter values, and App. \ref{app:open-loop} studies open-loop model performance.
A video of the experiments may be found online at \href{https://www.youtube.com/watch?v=RBUyvF6fzv0}{youtu.be/RBUyvF6fzv0}.

\section{Related Work}\label{sec:lit-review}
This work considers the problem of local trajectory planning and control,
    which is responsible for producing and tracking short-range trajectories that allow the vehicle to make progress towards a waypoint or goal.
In practice, this may be integrated as a component into a larger autonomy system by providing it the output of a global path planner
    \cite{shenEfficientGlobalTrajectory2024, krusiDrivingPointClouds2017}.

Safety is an important consideration for this problem as discussed in Sec. \ref{sec:intro}.
Another is dynamical feasibility: as the local trajectory planner produces plans, following them must be within the dynamic capabilities of the vehicle.

Several approaches have been considered for this problem domain.
State-lattice methods \cite{mcnaughtonMotionPlanningAutonomous2011, howardOptimalRoughTerrain2007}
    have been proposed for a variety of applications, including high-speed driving and extraterrestrial navigation,
but they are fundamentally incompatible with rugged off-road terrain or simply decompose the problem such that a solution to the current problem is still needed.
Potential field methods have also been explored for high-speed off-road driving \cite{shimodaPotentialFieldNavigation2005, iagnemmaOptimalNavigationHigh2008},
    but while the control algorithm is computationally efficient, it cannot guarantee dynamic feasibility nor safety.
Imitation learning approaches \cite{panImitationLearningAgile2020} have also been proposed,
    but they require an expert to learn from, so the planning problem still stands.

MPC \cite{williamsInformationTheoreticModelPredictive2018, yuNonlinearModelPredictive2021, hanModelPredictiveControl2024} is another approach to this problem.
It leverages a vehicle dynamics model to predict future states given a control input, and it uses a cost function to assign scalar costs to the associated trajectory prediction.
Planning and control is done iteratively, where each iteration
    constructs and solves an optimal control problem (OCP) to find the control input that minimizes cost given the vehicle's current state,
    then applies it to the vehicle.
While computationally more expensive than the aforementioned methods, it is capable of guaranteeing dynamic feasibility and safety.
Thus, this work considers MPC-based approaches.

In the prior work, most MPC-based approaches leverage some form of planar model (a.k.a., bicycle model)
    \cite{liuRoleModelFidelity2013, yuNonlinearModelPredictive2021, hanModelPredictiveControl2024}.
An advanced form of this model is the extended single-track model introduced in \cite{yuNonlinearModelPredictive2021}.
The differential equations still assume operation on a plane, but the model is applied to smooth non-planar terrain by creating a locally tangent approximation at each timestep.
Importantly, it requires that the terrain is locally planar at the length-scale of the chassis, and like the planar model, it assumes that neither tire liftoff nor suspension deflection occur.
Approaches using the extended single-track model have been demonstrated to be real-time feasible \cite{yuNonlinearModelPredictive2021}.
    Therefore, it is used as a \sota baseline in this work, and a full description is provided in App. \ref{app:models:3dof}.

A limitation of the extended single-track model is that it does not capture the vertical and suspension dynamics of the vehicle.
Other analytical models with additional dynamics exist in the literature \cite{eickNonlinearModelPredictive2015, shimUnderstandingLimitationsDifferent2007},
    but they have not been derived for use on non-planar terrain, making them inapplicable to the current problem.
Data-driven methods have also been used to model additional dynamics \cite{williamsInformationTheoreticModelPredictive2018, datarLearningModelPlan2023, leeLearningTerrainAwareKinodynamic2023}.
In particular, authors in \cite{leeLearningTerrainAwareKinodynamic2023} use a feed-forward neural network approach to model an SUV's dynamics in 3D space over arbitrary heightmap-encoded terrain.
However, the network uses two convolutional layers, one pooling layer, and five fully connected layers;
    it has nearly \num{60 000} parameters, so it is very likely that evaluating this model is much more computationally expensive than an explicit analytical model.
Since real-time feasibility is a key feature of a successful solution in this problem domain,
    and the approach given in \cite{leeLearningTerrainAwareKinodynamic2023} has not been demonstrated as such, it is not used as a baseline in this work.
However, the proposed analytical model contains similar dynamical properties.

To ensure vehicle safety, MPC formulations typically impose constraints on solutions to the OCP.
In the literature, rollover prevention constraints have been formulated to act on
    lateral acceleration \cite{liuRoleModelFidelity2013, hanModelPredictiveControl2024};
    steering angle \cite{liuNonlinearModelPredictive2018};
    chassis side-slip angle \cite{williamsInformationTheoreticModelPredictive2018};
    tire vertical load \cite{liuCombinedSpeedSteering2017, yuNonlinearModelPredictive2021};
    chassis roll and pitch \cite{leeLearningTerrainAwareKinodynamic2023};
    and terrain-interaction moments \cite{hanModelPredictiveControl2024, leeLearningTerrainAwareKinodynamic2023}.
Of these, chassis roll and pitch are most directly related to rollover (and pitch-over) prevention, but they are reliant on the model's ability to predict these states.
In the absence of direct prediction of these states, such as with a planar model, tire vertical load and lateral acceleration are the next most directly related to rollover prevention:
    tire liftoff is a necessary condition for rollover, so constraining tire loads to remain positive precludes vehicle rollover.

There is a strong relationship between the extended single-track model's predicted tire vertical load and lateral acceleration (discussed in Sec \ref{sec:modeling-safety:analysis}).
    Thus, the lateral-acceleration constraint is used as a \sota baseline in this work, but the findings are shown to be generally applicable to the tire-liftoff constraint as well.
The limitation of such a constraint is that it only considers rollovers caused by tire forces within the lateral-longitudinal plane of the vehicle.
It neglects rollovers caused by terrain roughness and out-of-plane dynamics, leaving the vehicle unprotected against them (discussed in Sec \ref{sec:modeling-safety:discussion}).
As these factors are significant in off-road driving, neglecting them puts the the autonomous vehicle at risk.

Outside of applications to MPC safety constraint formulation, several other metrics related to rollover are proposed in the literature.
Perhaps the most simple is static stability factor (SSF), which measures geometric properties of the vehicle and considers no dynamic terms \cite{petersAnalysisRolloverStability2006}.
It may be used to compare the intrinsic stability of different vehicles, but is not useful as a real-time metric of stability.
Load transfer ratio (LTR) considers the distribution of tire normal force along the vehicle's lateral axis \cite{petersAnalysisRolloverStability2006}.
This metric is similar to the previously considered tire-vertical-load and lateral-acceleration constraints and suffers the same problems when used with a planar model.
In addition, it is overconservative for off-road applications since it is possible to experience tire liftoff without rollover.

The force-angle \cite{papadopoulosNewMeasureTipover1996, papadopoulosForceAngleMeasureTipover2000, petersAnalysisRolloverStability2006}
and energy-stability \cite{messuriAutomaticBodyRegulation1985, ghasempoorMeasureMachineStability1995, naleczInvestigationDynamicMeasures1993} families of metrics also exist.
Both consider the support polygon of the vehicle, meaning a quadrilateral formed by the four tires.
The force angle measure (FAM) considers the vector of summed forces (gravitational and d'Alambert's) acting on the chassis and computes its angle relative to the support polygon.
When forces are directed inside the polygon, the vehicle is stable and the metric is positive.
The metric decreases to zero and becomes increasingly negative as the vector rotates to point at the perimeter then finally outside.
Importantly, a negative value of this metric does not indicate rollover is occurring or will occur; it indicates that the current forces, if held constant, will eventually produce rollover.
A form of this metric has been applied to high-speed mobile robotics \cite{petersAnalysisRolloverStability2006},
    but because aggressive maneuvering may require subjecting the vehicle to brief destabilizing forces that do not persist long enough to cause rollover,
    the above property makes it overconservative for high-speed off-road applications.

The latter family of metrics, energy-stability, consider the location of the chassis center of mass (CoM) relative to the support polygon.
The energy stability margin (ESM) measures the minimum potential energy difference between the system's current state and an unstable equilibrium reached by rotating the vehicle about an edge of the support polygon, such that the CoM lies directly above the line of contact.
In other words, it measures the minimum required potential energy change to bring the system to rollover or pitch-over.
A negative value of this metric indicates that rollover or pitch-over is actively occurring, and it may continue to grow in the negative direction as rollover or pitch-over continues.
Thus, ESM does not suffer from the same overconservativeness as FAM and is therefore preferred in this work.
Notably, implementing this metric requires a vehicle model capable of accurately predicting roll and pitch, something the extended single-track model is not designed for.
As the vehicle model proposed in this work has no such limitation, ESM is used to formulate the rollover-prevention constraint of the proposed approach.
To the best of the authors' knowledge, this is the first application of this metric to constraint design in local trajectory planning for high-speed autonomous mobile robotics applications.

Equally important to the formulation, MPC approaches require a means to solve the OCP in real time.
Prior studies have used both gradient-based \cite{yuNonlinearModelPredictive2021} and sampling-based \cite{williamsInformationTheoreticModelPredictive2018} methods.
In addition to not requiring gradients, the latter can accelerate drawing samples, simulating trajectories, and computing costs through the use of GPGPU (General-Purpose computing on Graphics Processing Units) programming.
For example, authors in \cite{hanModelPredictiveControl2024} use the model predictive path integral control (MPPI) algorithm \cite{williamsAggressiveDrivingModel2016}, a specific type of sampling-based method, to demonstrate operation at \qty{33.8}{\Hz} with an NVIDIA 3080 desktop GPU.
Because the proposed vehicle model is not analytically differentiable, this work uses a sampling-based method,
    and GPGPU programming is leveraged to demonstrate real-time feasibility with an NVIDIA mobile GPU.

In summary, current methods lack the ability to plan and execute the extreme maneuvers required to autonomously navigate rugged off-road terrain while simultaneously ensuring vehicle safety.
Approaches using planar vehicle models may be real-time feasible, but they lack the ability to predict and mitigate an entire class of rollovers,
    forcing one to compromise on either safety, speed, or terrain ruggedness.
This can be overcome by using a vehicle model with additional dynamics, but these approaches have not yet been demonstrated as real-time feasible and lack thorough analysis in realistic off-road environments.
This work proposes an analytical vehicle model with the necessary dynamics and demonstrates real-time feasibility via GPGPU computing.
The performance of the approach is rigorously evaluated through theoretical and empirical means, and data from both simulated trials and physical experiments are presented.

\section{State-of-the-Art Vehicle and Safety Constraint Models}\label{sec:modeling-safety}
\subsection{Extended Single-Track Vehicle Dynamics Model}\label{app:models:3dof}

\begin{figure}[ht]
    \centering
    \includegraphics[width=0.9\linewidth]{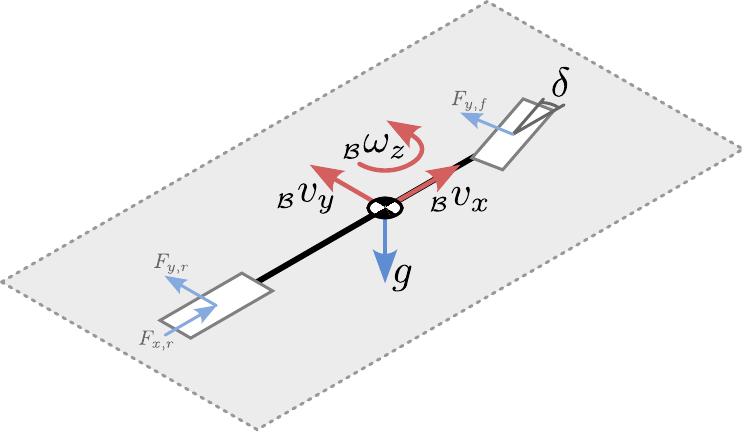}
    \caption{The extended single-track vehicle dynamics model, a \sota baseline}
    \label{fig:models:3dof}
\end{figure}

The extended single-track model, diagrammed in Fig. \ref{fig:models:3dof},  is a single-track, planar vehicle model with prescribed longitudinal velocity rate and steering rate.
It is applied to non-planar terrain by continually approximating the terrain as a plane locally tangential to the surface under the center of mass (CoM).
Due to the planar terrain approximation, it is only valid on smooth terrain.
The terrain is also assumed to be rigid and expressed as a function of location, but
this function need not be analytical: a heightmap is sufficient so long as numerical first derivatives can be taken to evaluate slope.

The model was introduced in \cite{yuNonlinearModelPredictive2021}, and the version leveraged in this work is identical except for the tire model.
The model is reproduced here to remove any ambiguity; this reporting
    corrects minor errors,
    expresses assumptions more thoroughly,
    uses a notation consistent with the model proposed in Sec. \ref{app:models:6dof},
    and provides several necessary additional equations.

The model makes the following assumptions:
\begin{enumerate}[\hspace{5pt}{A}1)]
    \item The terrain is rigid, and the terrain height can be determined for any $(x,y)$ location
    \item \label{app:models:3dof:assumptions:planar}
        The terrain is sufficiently smooth to be locally planar at the length-scale of the chassis
    \item \label{app:models:3dof:assumptions:liftoff}
        The tires are always in contact with the terrain
    \item \label{app:models:3dof:assumptions:suspension}
        The suspension deflection is negligible:
        the vehicle's CoM is a constant distance from the local plane, and the chassis' vertical axis is always normal to the terrain
    \item \label{app:models:3dof:assumptions:angvel}
        The longitudinal and lateral components of angular velocity are negligible
    \item The longitudinal velocity of the chassis is positive and may have a prescribed rate of change
\end{enumerate}

The state and control vectors are defined in \cite{yuNonlinearModelPredictive2021} as
\begin{equation}
    \bm{\xi} \triangleq \begin{bmatrix}
        x \\ y \\ \psi \\ \resveccomp{v}{B}{y} \\ \resveccomp{\omega}{B}{z} \\ \delta \\ \resveccomp{v}{B}{x}
    \end{bmatrix} = \begin{bmatrix}
        \text{CoM $x$ position} \\
        \text{CoM $y$ position} \\
        \text{chassis yaw angle} \\
        \text{CoM lateral velocity} \\
        \text{chassis angular velocity about $\mathbf{b}_x$} \\
        \text{steering angle} \\
        \text{CoM longitudinal velocity} \\
    \end{bmatrix}
    \label{eq:models:3dof:state}
\end{equation}
and
\begin{equation}
    \bm{\zeta} \triangleq \begin{bmatrix}
        \dot{\delta} \\ \dot{\resveccomp{v}{B}{x}}
    \end{bmatrix} = \begin{bmatrix}
        \text{steering rate} \\
        \text{CoM longitudinal velocity rate} \\
    \end{bmatrix}
    \label{eq:models:3dof:control}
\end{equation}

The derivative of the state is computed using
\begin{equation}
    \dot{\bm{\xi}} = A(\bm{\xi}) + B \bm{\zeta}
    \label{eq:models:3dof:update}
\end{equation}
where
\begin{equation}
    A(\bm{\xi}) = \begin{bmatrix}
        \resveccomp{v}{W}{x} \\[4pt]
        \resveccomp{v}{W}{y} \\[4pt]
        \resveccomp{\omega}{B}{z} \\[4pt]
        \frac{F_{y,f} + F_{y,r}}{M} + \resveccomp{g}{B}{y} - \resveccomp{\omega}{B}{z} \resveccomp{v}{B}{x}\\[4pt]
        \frac{F_{y,f} L_f \cos(\delta) - F_{y,r} L_r}{J_{zz}} \\[4pt]
        0 \\[4pt]
        0
    \end{bmatrix}
\label{eq:models:3dof:A}
\end{equation}
and
\begin{equation}
    B = \begin{bmatrix}
        0 & 0 & 0 & 0 & 0 & 1 & 0 \\
        0 & 0 & 0 & 0 & 0 & 0 & 1
    \end{bmatrix}^T
    \label{eq:model:B}
\end{equation}
as in \cite{yuNonlinearModelPredictive2021},
where $F_{y,f}$ and $F_{y,r}$ are the lateral forces on the front and rear virtual tires, respectively,
$\bm{g}$ is the gravity vector,
and other parameters are as defined in App. \ref{app:notation}.

Notably, the vehicle experiences roll, pitch, and heave dynamics, but these terms do not appear in the state vector.
Due to assumption A\ref{app:models:3dof:assumptions:suspension}, these dynamics are passively induced by the terrain as the vehicle moves between local planes.

To simplify the computation necessary to recover these dynamics from the state vector and local plane, the model utilizes Euler angles following the 1-2-3 convention.
That is, $\mathcal{B}$ is generated from $\mathcal{W}$ by first rotating along $\mathbf{w}_x$ by $\phi$,
then along the new $y$ axis by $\theta$;
then along the new $z$ axis, $\mathbf{b}_z$ by $\psi$.
An arbitrary vector, $\bm{q}$, is transformed via
\begin{equation}
\resvec{q}{W}
= \begin{bmatrix}
    c_\theta c_\psi & -c_\theta s_\psi & s_\theta \\
    c_\phi s_\psi + c_\psi s_\phi s_\theta & c_\phi c_\psi - s_\phi s_\theta s_\psi & -c_\theta s_\phi \\
    s_\phi s_\psi - c_\phi c_\psi s_\theta & c_\psi s_\phi + c_\phi s_\theta s_\psi & c_\phi c_\theta \\
\end{bmatrix}
\resvec{q}{B}
\label{app:models:3dof:rotation}
\end{equation}
where $c_\bullet$ and $s_\bullet$ are shorthand for $\cos(\bullet)$ and $\sin(\bullet)$.

Under this convention, the roll and pitch angles are determined by the normal vector of the terrain under the vehicle's CoM.
Essentially, the roll and pitch define the local plane, and the yaw rotates within the it.
The relationship, from \cite{yuNonlinearModelPredictive2021}, is
\begin{equation}
    \resvec{\tau}{W} = \begin{bmatrix}
    \sin(\theta) \\
    -\cos(\theta) \sin(\phi) \\
    \cos(\phi) \cos(\theta)
\end{bmatrix}
\end{equation}
where $\bm{\tau}$ is the unit-length normal vector, found by evaluating the slope of the terrain at the position of the vehicle's CoM:
\begin{equation}
\resvec{\tau}{W} \propto
\begin{bmatrix}
    f_x(x,y) & f_y(x,y) & 1
\end{bmatrix}^T
\label{eq:models:3dof:tau2}
\end{equation}
where $f_x$ and $f_y$ are partial derivatives of $f$, the terrain height function,
taken with respect to $x$ and $y$, respectively.

Additionally, the following are true via assumption A\ref{app:models:3dof:assumptions:suspension}, assumption A\ref{app:models:3dof:assumptions:angvel}, and the 1-2-3 Euler angle convention:
\begin{align}
    z &= f(x,y) + (h+R) \mathbf{b}_z \\
    \resvec{v}{B} &= \begin{bmatrix} \resveccomp{v}{B}{x} & \resveccomp{v}{B}{y} & 0\end{bmatrix}^T \\
    \resvec{\omega}{B} &= \begin{bmatrix} 0 & 0 & \resveccomp{\omega}{B}{z}\end{bmatrix}^T
\end{align}
where $z$ denotes the $z$ position of the CoM.
Correcting from \cite{yuNonlinearModelPredictive2021}, the rate of change of the yaw angle, $\dot{\psi}$, is simply $\resveccomp{\omega}{B}{z}$.

The lateral tire forces, $F_{y,f}$ and $F_{y,r}$, are computed
using a reduced-order tire model.
The model was created by fitting to synthetic data produced by the data-driven tire model used in \cite{yuRealTimeTerrainAdaptiveLocal2025}.
It utilizes a sigmoid function:
\begin{equation}\label{eq:models:tire}
    \frac{F_{y,i}}{F_{z,i}} = \frac{- C_i \alpha_i \mu_i}{\sqrt{\mu_i^2 + (C_i \alpha_i)^2}}
\end{equation}
where $i$ indexes the tire under analysis,
$\alpha_i$ is the slip angle,
$F_{z,i}$ is the normal force,
$C_i$ is the cornering stiffness in the linear region,
and $\mu_i$ is the maximum value of $\lvert \frac{F_{y,i}}{F_{z,i}} \rvert$.

The extended single-track model evaluates \eqref{eq:models:tire}
on the front and rear virtual tires. Their slip angles are computed as
\begin{subequations}
\begin{align}
    \alpha_f &= \arctan \left( \frac{\resveccomp{v}{B}{y} + \resveccomp{\omega}{B}{z} L_f}{\resveccomp{v}{B}{x}} \right) - \delta \\
    \alpha_r &= \arctan \left( \frac{\resveccomp{v}{B}{y} - \resveccomp{\omega}{B}{z} L_r}{\resveccomp{v}{B}{x}} \right)
\end{align}
\end{subequations}

Normal forces for each physical tire,
which may be used to verify assumption A\ref{app:models:3dof:assumptions:liftoff} and/or enforce constraints,
are computed via the following corrected and simplified relations from \cite{yuNonlinearModelPredictive2021}:
\begin{subequations}
\label{eq:models:3dof:Fz_all}
\begin{align}
    F_{z,fl} &= ( -K_{zz,f} \, \resveccomp{g}{B}{z} - K_{zx} \, \resveccomp{a}{B}{x}) \left( \frac{1}{2} - \frac{K_{zy} \, \resveccomp{a}{B}{y}}{-M \resveccomp{g}{B}{z}} \right) \\
    F_{z,fr} &= ( -K_{zz,f} \, \resveccomp{g}{B}{z} - K_{zx} \, \resveccomp{a}{B}{x}) \left( \frac{1}{2} + \frac{K_{zy} \, \resveccomp{a}{B}{y}}{-M \resveccomp{g}{B}{z}} \right) \\
    F_{z,rl} &= ( -K_{zz,r} \, \resveccomp{g}{B}{z} + K_{zx} \, \resveccomp{a}{B}{x}) \left( \frac{1}{2} - \frac{K_{zy} \, \resveccomp{a}{B}{y}}{-M \resveccomp{g}{B}{z}} \right) \\
    F_{z,rr} &= ( -K_{zz,r} \, \resveccomp{g}{B}{z} + K_{zx} \, \resveccomp{a}{B}{x}) \left( \frac{1}{2} + \frac{K_{zy} \, \resveccomp{a}{B}{y}}{-M \resveccomp{g}{B}{z}} \right)
\end{align}
\end{subequations}
where $K_{zz,f}$, $K_{zz,r}$, $K_{zx}$, and $K_{zy}$ are the front-vertical, rear-vertical, longitudinal, and lateral load transfer coefficients, respectively, given by
\begin{align}
    K_{zz,f} &\triangleq M \frac{L_r}{L_f+L_r} \label{eq:models:3dof:Kzzf} \\
    K_{zz,r} &\triangleq M \frac{L_f}{L_f+L_r} \label{eq:models:3dof:Kzzr} \\
    K_{zx} &\triangleq M \frac{h+R}{L_f+L_r} \label{eq:models:3dof:Kzx} \\
    K_{zy} &\triangleq M \frac{h+R}{e} \label{eq:models:3dof:Kzy}
\end{align}
and $\resveccomp{a}{B}{x}$ and $\resveccomp{a}{B}{y}$ are the longitudinal and lateral components of the vehicle CoM's acceleration, respectively, given by
\begin{align}
\resveccomp{a}{B}{x} = \resveccomp{\dot{v}}{B}{x} - \resveccomp{\omega}{B}{z} \resveccomp{v}{B}{y} \label{eq:models:3dof:abx} \\
\resveccomp{a}{B}{y} = \resveccomp{\dot{v}}{B}{y} + \resveccomp{\omega}{B}{z} \resveccomp{v}{B}{x} \label{eq:models:3dof:aby}
\end{align}

The normal forces for the virtual tires, needed to evaluate \eqref{eq:models:tire}, are found by summing the forces from \eqref{eq:models:3dof:Fz_all}:
\begin{subequations}
\label{eq:models:3dof:Fz_lon}
\begin{align}
    F_{z,f} &= -K_{zz,f} \, \resveccomp{g}{B}{z} - K_{zx} \, \resveccomp{a}{B}{x} \\
    F_{z,r} &= -K_{zz,r} \, \resveccomp{g}{B}{z} + K_{zx} \, \resveccomp{a}{B}{x}
\end{align}
\end{subequations}

\subsection{Analysis} \label{sec:modeling-safety:analysis}
The purpose of the following analysis is to establish a fundamental shortcoming of \sota approaches.
It proves that the typically applied rollover prevention constraints do
not protect against rollovers caused by terrain roughness and/or suspension dynamics.
This means that protection against these types of rollovers is left up to chance,
and it motivates the development of the proposed formulation in Sec. \ref{sec:formulation}.

The derivation takes place in three parts.
First, Sec. \ref{sec:modeling-safety:analysis:equivalence} demonstrates how a tire-liftoff constraint
may be reformulated as an equivalent lateral-acceleration constraint.
Notably, this justifies the use of a lateral-acceleration constraint as a \sota baseline in this work.
Next, Sec. \ref{sec:modeling-safety:analysis:violation} demonstrates how the possibility of violating
the lateral-acceleration (and corresponding tire-liftoff) constraint is governed by the traction of the terrain.
Lastly, Sec. \ref{sec:modeling-safety:analysis:proof} proves how this results in the aforementioned shortcoming,
a lack of protection against certain types of rollovers, when applied to the problem domain studied in this work.

The analysis considers a tire-liftoff constraint, which, as reviewed in Sec. \ref{sec:lit-review},
is a \sota approach for ensuring the safety of an autonomous off-road vehicle
against rollover.
A scenario where the chassis experiences negligible longitudinal acceleration is considered,
    which, as seen in \eqref{eq:models:3dof:abx},
    occurs when a constant longitudinal velocity is prescribed and the yaw rate and lateral velocity are both small enough that their product is approximately zero.

\subsubsection{Constraint Equivalencies} \label{sec:modeling-safety:analysis:equivalence}
The tire-liftoff constraint asserts that the tire normal forces given in \eqref{eq:models:3dof:Fz_all} are all strictly positive,
meaning $F_{z,i} > 0 \quad \forall \,i \in \{fl,fr,rl,rr\}$.
Under the assumed condition of negligible longitudinal acceleration, the tire normal forces simplify as
\begin{subequations}
    \label{eq:modeling-safety:Fz_all}
\begin{align}
    F_{z,fl}^\prime &= \frac{K_{zz,f}}{M} \left( \frac{-M \resveccomp{g}{B}{z}}{2} - K_{zy} \, \resveccomp{a}{B}{y} \right) \\
    F_{z,fr}^\prime &= \frac{K_{zz,f}}{M} \left( \frac{-M \resveccomp{g}{B}{z}}{2} + K_{zy} \, \resveccomp{a}{B}{y} \right) \\
    F_{z,rl}^\prime &= \frac{K_{zz,r}}{M} \left( \frac{-M \resveccomp{g}{B}{z}}{2} - K_{zy} \, \resveccomp{a}{B}{y} \right) \\
    F_{z,rr}^\prime &= \frac{K_{zz,r}}{M} \left( \frac{-M \resveccomp{g}{B}{z}}{2} + K_{zy} \, \resveccomp{a}{B}{y} \right)
\end{align}
\end{subequations}
where it is noted that $\resveccomp{g}{B}{z}$ is a negative quantity.

Due to the symmetry of \eqref{eq:modeling-safety:Fz_all}, the tire-liftoff constraint may be equivalently expressed as
\begin{equation}
    \label{eq:modeling-safety:liftoff_ineq1}
    \lvert \resveccomp{a}{B}{y} \rvert < \frac{-M \resveccomp{g}{B}{z}}{2 K_{zy}}
\end{equation}

The load transfer equations given in \eqref{eq:models:3dof:Fz_all} and simplified in \eqref{eq:modeling-safety:Fz_all} are originally from \cite{doumiatiLateralLoadTransfer2009},
    which did not consider non-horizontal surfaces in its derivation.
Due to this, the in-plane components of the gravity vector are neglected.
For the terrain considered in this study, which does not include extreme slopes, this assumption is justifiable.
However, to assist with analytical simplification in this analysis, this assumption is removed by subtracting the missing gravity components from chassis acceleration.
Thus, \eqref{eq:modeling-safety:liftoff_ineq1} is corrected as
\begin{equation}
    \label{eq:modeling-safety:liftoff_ineq2}
    \lvert \resveccomp{a}{B}{y} - \resveccomp{g}{B}{y} \rvert < \frac{-M \resveccomp{g}{B}{z}}{2 K_{zy}}
\end{equation}
where $\resveccomp{g}{B}{y}$ is the lateral component of the gravity vector.
By dividing both sides by $-\resveccomp{g}{B}{z}$, the right-hand side of the inequality is made to contain only fixed vehicle properties:
\begin{equation}
    \label{eq:modeling-safety:liftoff_ineq3}
    \frac{\lvert \resveccomp{a}{B}{y} - \resveccomp{g}{B}{y} \rvert}{- \resveccomp{g}{B}{z}} < \frac{M }{2 K_{zy}}
\end{equation}
This inequality expresses a lateral-acceleration constraint.
Thus, a tire-liftoff constraint may be equivalently reformulated as a lateral-acceleration constraint when the longitudinal acceleration is negligible.
The constraint may be expressed as
\begin{equation}
    \label{eq:modeling-safety:aby_ineq}
    \frac{\lvert \resveccomp{a}{B}{y} - \resveccomp{g}{B}{y} \rvert}{- \resveccomp{g}{B}{z}} < \frac{ \resveccomp{\bar{a}}{B}{y} }{g}
\end{equation}
where $\resveccomp{\bar{a}}{B}{y}$ is the critical lateral-acceleration boundary for tire liftoff on flat terrain.
It may be found empirically or via
\begin{equation}
    \label{eq:modeling-safety:aby_bar_def}
    \resveccomp{\bar{a}}{B}{y} \triangleq \frac{M g}{2 K_{zy}}
\end{equation}

\subsubsection{Possibility of Constraint Violation} \label{sec:modeling-safety:analysis:violation}
To compute the left-hand side of \eqref{eq:modeling-safety:aby_ineq}, \eqref{eq:models:3dof:A} is first substituted into \eqref{eq:models:3dof:aby}, yielding
\begin{equation}
    \label{eq:deficit:aby}
    \resveccomp{a}{B}{y} = \frac{F_{y,f} + F_{y,r}}{M} + \resveccomp{g}{B}{y}
\end{equation}
where $F_{y,f}$ and $F_{y,r}$ are the lateral tire forces on the front and rear virtual tires, respectively.
An important property of the tire model is that the maximum lateral force it may predict is bounded by the tire normal force and terrain traction.
The maximum lateral force for virtual tire $i \in \{f, r\}$ is constrained as follows:
\begin{equation}
    \label{eq:deficit:Fy}
    \lvert F_{y,i} \rvert \leq \mu F_{z,i}^\prime
\end{equation}
where $\mu$ is a model parameter denoting the maximum coefficient of friction supported by the terrain.
Virtual tire normal forces are found by laterally summing the physical tire normal forces.
Performing this operation on \eqref{eq:modeling-safety:Fz_all} yields
\begin{equation}
    \label{eq:deficit:Fz}
    F_{z,i}^\prime = -K_{zz,i} \, \resveccomp{g}{B}{z}
\end{equation}
Subtracting $\resveccomp{g}{B}{y}$ from both sides of \eqref{eq:deficit:aby} and
performing substitution via \eqref{eq:deficit:Fy} and \eqref{eq:deficit:Fz} yields an inequality constraining the maximum lateral acceleration that the model may predict:
\begin{equation}
    \label{eq:deficit:aby_sub}
    \lvert \resveccomp{a}{B}{y} - \resveccomp{g}{B}{y} \rvert \leq - \resveccomp{g}{B}{z} \, \mu \frac{K_{zz,f} + K_{zz,r} }{M}
\end{equation}
The fraction in \eqref{eq:deficit:aby_sub} simplifies to 1, and both sides may be divided by $-\resveccomp{g}{B}{z}$ to express the maximum value of the left-hand side of \eqref{eq:modeling-safety:aby_ineq}:
\begin{equation}
    \label{eq:modeling-safety:aby_ineq_rhs_star}
    \frac{\lvert \resveccomp{a}{B}{y} - \resveccomp{g}{B}{y} \rvert}{- \resveccomp{g}{B}{z}} \leq \mu
\end{equation}
Thus, the possibility of violating \eqref{eq:modeling-safety:aby_ineq} is governed by
\begin{equation}
    \label{eq:modeling-safety:mu_ineq}
    \mu \geq \frac{ \resveccomp{\bar{a}}{B}{y} }{g}
\end{equation}
If this inequality evaluates false, it is not possible to violate the lateral-acceleration (and equivalent tire-liftoff) constraint with any state or control vector that maintains the assumed negligible longitudinal acceleration.

\subsubsection{Application to Off-Road Domains} \label{sec:modeling-safety:analysis:proof}
The previous findings are relevant to applications that meet the following two conditions:
First, the terrain traction is low enough, and the vehicle is stable enough, that \eqref{eq:modeling-safety:mu_ineq} evaluates false.
Second, it is still possible to roll the vehicle due to the effects of non-planar geometry,
even without experiencing non-negligible longitudinal acceleration.
Both of these conditions are true for the vehicle and scenarios considered in this work.

The relevance is demonstrated through a simple proof by contradiction.
The proof supposes that a tire-liftoff constraint implemented with the extended single-track model \emph{is} capable of protecting the vehicle against all types of rollovers.
Due to the first condition, the equivalent lateral-acceleration constraint is impossible to violate.
This implies the vehicle cannot be rolled so long as longitudinal acceleration is negligible,
contradicting the second condition.
Thus, the initial premise is false, meaning the tire-liftoff constraint \emph{is not} capable of protecting the vehicle against all types of rollovers.

To summarize, being unable to violate the safety constraint does not guarantee the vehicle is protected from rollover.
Even on low-traction terrains, it is still possible to roll the vehicle due to the effects of terrain roughness and/or suspension dynamics.
This proves that a tire-liftoff or lateral-acceleration constraint, implemented with the extended single-track model,
    is insufficient to protect autonomous off-road vehicles from rollover when navigating rough terrain at high-speed.

\subsection{Discussion} \label{sec:modeling-safety:discussion}
The efficacy of a safety constraint depends on the model's ability to predict the terms included in the constraint.
In the case of the \sota approach analyzed in the previous section, it depends on the extended single-track model's ability to predict tire loading.

Due to the extended single-track model's simplicity, its ability to predict tire loading is limited:
    the model lacks roll, pitch, and heave dynamics, and it assumes the terrain is locally planar, meaning it ignores the unique geometry under each tire.
As such, the only tire loading dynamics it predicts are those caused by longitudinal and lateral acceleration.
This is demonstrated by the re-expression of a tire-liftoff constraint as a lateral-acceleration constraint in the previous analysis.

Therefore, when the extended single-track model is used to evaluate a tire-liftoff constraint, it only evaluates tire liftoff caused by load transfer due to in-plane acceleration.
It does not consider tire liftoff caused by vertical dynamics or rough terrain, since the model is not designed to predict these phenomena.
Thus, the safety constraint protects against only one type of tire liftoff, and the vehicle is left vulnerable to the others.
To safely use this approach, these vulnerabilities must be mitigated through alternate means.

A straightforward way to achieve this is by reducing operational domain via simple constraints on speed and location.
To guarantee safety, the domain must be restricted such that the vehicle cannot experience rollover via vertical dynamics or rough terrain.
This may take the form of no-go areas around rough terrain, or significant speed restrictions in these regions, such that the suspension dynamics are quasi-static.
However, this would significantly reduce the mobility of the vehicle.

To leverage the full limits of mobility of an off-road vehicle platform, it must be enabled to safely operate in domains with a risk of rollover due to rough terrain and vertical dynamics.
The \sota approach cannot achieve this, since it lacks sufficient modeling fidelity to provide complete safety against rollover via a tire-liftoff constraint.
Therefore, a successful approach must include mechanisms to comprehensively predict and mitigate a rollovers across a larger operational domain.
The next section proposes such an approach.

\section{Proposed MPC Formulation}\label{sec:formulation}

\subsection{Single Rigid Body Vehicle Dynamics Model}\label{app:models:6dof}

\begin{figure}[ht]
    \centering
    \includegraphics[width=0.9\linewidth]{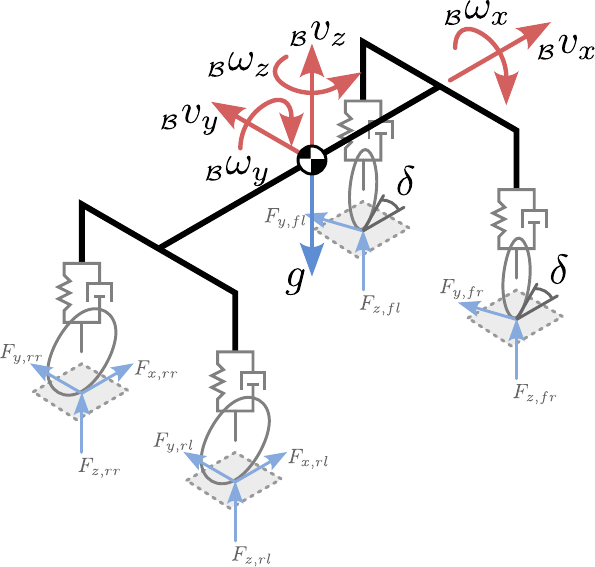}
    \caption{The single rigid body vehicle dynamics model, a higher-fidelity alternative to the extended single-track model}
    \label{fig:models:6dof}
\end{figure}

To predict and mitigate rollovers more comprehensively than the \sota approach, a higher fidelity model is used to predict vehicle dynamics.
This model, denoted as the single rigid body model,
is derived with the aim of having the minimal complexity required to model rollover caused by rough terrain and vertical motion.

The single rigid body model, diagrammed in Fig. \ref{fig:models:6dof}, is a 4-wheeled vehicle model that treats the vehicle as a single rigid body in 3D space, with the lumped mass and inertia properties of the whole vehicle.
Due to this, the roll, pitch, and heave dynamics are independent degrees of freedom, distinct from the terrain.
It models the suspension as independent linear spring-dampers at each wheel, and the wheels are considered massless.
Additionally, the tires are not assumed to drive on the same local plane, nor are they required to be in constant contact with the terrain.
This relaxes the smoothness requirement compared to the extended single-track model, enabling it to be simulated over rough surfaces.

The model makes the following assumptions:
\begin{enumerate}[\hspace{05pt}{A}1)]
    \item The terrain is rigid, and the terrain height can be determined for any $(x,y)$ location
    \item \label{app:models:6dof:assumptions:planar}
        The terrain is sufficiently smooth to be locally planar at the length-scale of a wheel
    \item The suspension deflection occurs in the chassis vertical direction,
        and the contact patch is directly below the wheel hub in the chassis frame
    \item The location at which the tire forces act is constant in the chassis frame, despite suspension deflection
    \item The wheels have negligible mass
    \item \label{app:models:6dof:assumptions:longvel}
        The longitudinal velocity of the chassis is positive and may have a prescribed rate of change
\end{enumerate}

The state vector is defined as
\begin{equation}
    \bm{\xi} \triangleq \begin{bmatrix}
        x \\ y \\ z \\
        \psi \\ \theta \\ \phi \\
        \resveccomp{v}{B}{x} \\ \resveccomp{v}{B}{y} \\ \resveccomp{v}{B}{z} \\
        \resveccomp{\omega}{B}{x} \\ \resveccomp{\omega}{B}{y} \\ \resveccomp{\omega}{B}{z} \\
        \delta
    \end{bmatrix} = \begin{bmatrix}
        \text{CoM $x$ position} \\
        \text{CoM $y$ position} \\
        \text{CoM $z$ position} \\
        \text{chassis yaw angle} \\
        \text{chassis pitch angle} \\
        \text{chassis roll angle} \\
        \text{CoM longitudinal velocity} \\
        \text{CoM lateral velocity} \\
        \text{CoM vertical velocity} \\
        \text{chassis angular velocity about $\mathbf{b}_x$} \\
        \text{chassis angular velocity about $\mathbf{b}_y$} \\
        \text{chassis angular velocity about $\mathbf{b}_z$} \\
        \text{steering angle}
    \end{bmatrix}
    \label{eq:models:6dof:state}
\end{equation}
and the control vector is identical to \eqref{eq:models:3dof:control}.

The state update equation is also identical to \eqref{eq:models:3dof:update}, where
\begin{equation}
    A(\bm{\xi}) = \begin{bmatrix}
        \resveccomp{v}{W}{x} \\[4pt]
        \resveccomp{v}{W}{y} \\[4pt]
        \resveccomp{v}{W}{z} \\[4pt]
        \dot{\psi} \\[4pt]
        \dot{\theta} \\[4pt]
        \dot{\phi} \\[4pt]
        0 \\[4pt]
        \frac{F_y}{M} + \resveccomp{g}{B}{y}
        + \resveccomp{\omega}{B}{x} \resveccomp{v}{B}{z}
        - \resveccomp{\omega}{B}{z} \resveccomp{v}{B}{x} \\[4pt]
        \frac{F_z}{M} + \resveccomp{g}{B}{z}
        - \resveccomp{\omega}{B}{x} \resveccomp{v}{B}{y}
        + \resveccomp{\omega}{B}{y} \resveccomp{v}{B}{x} \\[4pt]
        \frac{M_x + (J_{yy}-J_{zz}) \resveccomp{\omega}{B}{y} \resveccomp{\omega}{B}{z}}{J_{xx}}\\[4pt]
        \frac{M_y - (J_{xx}-J_{zz}) \resveccomp{\omega}{B}{x} \resveccomp{\omega}{B}{z}}{J_{yy}}\\[4pt]
        \frac{M_z + (J_{xx}-J_{yy}) \resveccomp{\omega}{B}{x} \resveccomp{\omega}{B}{y}}{J_{zz}}\\[4pt]
        0
    \end{bmatrix}
    \label{eq:models:6dof:A}
\end{equation}
and
\begin{equation}
    B = \begin{bmatrix}
        0_{2 \mathrm{x} 6} &
        \begin{aligned} 0 \\ 1 \end{aligned} &
        0_{2 \mathrm{x} 5} &
        \begin{aligned} 1 \\ 0 \end{aligned}
    \end{bmatrix}^T
    \label{eq:6dof:B}
\end{equation}
$F_y$, $F_z$, $M_x$, $M_y$, and $M_z$ are summed forces and moments computed by separate equations described in this section.

The implementation of this model utilizes Euler angles following the 3-2-1 convention.
That is, $\mathcal{B}$ is generated from $\mathcal{W}$ by first rotating along $\mathbf{w}_z$ by $\psi$,
then along the new $y$ axis by $\theta$,
then along the new $x$ axis, $\mathbf{b}_x$, by $\phi$.
An arbitrary vector, $\bm{q}$, is transformed via
\begin{equation}
    \resvec{q}{W} = \begin{bmatrix}
        c_\psi c_\theta & c_\psi s_\phi s_\theta - c_\phi s_\psi & s_\phi s_\psi + c_\phi c_\psi s_\theta \\
        c_\theta s_\psi & c_\phi c_\psi + s_\phi s_\psi s_\theta & c_\phi s_\psi s_\theta - c_\psi s_\phi \\
        -s_\theta & c_\theta s_\phi & c_\phi c_\theta \\
    \end{bmatrix} \resvec{q}{B}
    \label{eq:models:6dof:rotation-matrix}
\end{equation}
where $c_\bullet$ and $s_\bullet$ are shorthand for $\cos(\bullet)$ and $\sin(\bullet)$.

This rotation representation is selected out of preference, and it is not intrinsically required by the model.
Notably, it experiences gimbal lock when $\theta = \frac{\pi}{2}$.
However, no numerical issues are observed due to this, as this situation corresponds to the vehicle's forward axis,
$\mathbf{b}_x$, pointing directly upwards along the world's vertical axis, $\mathbf{w}_z$.
The vehicle's dynamics do not approach this region, so it is not a concern for applications of this work.

Under this convention, Euler angle rates can be determined from angular velocity via
\begin{equation}
    \begin{bmatrix}
        \dot{\psi} \\ \dot{\theta} \\ \dot{\phi}
    \end{bmatrix} = \begin{bmatrix}
        0 & \frac{\sin(\phi)}{\cos(\theta)} & \frac{\cos(\phi)}{\cos(\theta)} \\
        0 & \cos(\phi) & -\sin(\phi) \\
        1 & \frac{\sin(\phi) \sin(\theta)}{\cos(\theta)} & \frac{\cos(\phi) \sin(\theta)}{\cos(\theta)}
    \end{bmatrix} \resvec{\omega}{B}
    \label{eq:models:6dof:euler-rates}
\end{equation}

Several steps are required to compute the forces and moments occurring at each wheel.
The procedure for wheel $i \in \{fl,\ fr,\ rl,\ rr\}$
    (corresponding to front-left, front-right, rear-left, and rear-right, respectively)
    is as follows:

First, the position and velocity of the chassis, at a virtual point coincident with the nominal contact patch is found via
\begin{align}
    \label{eq:models:6dof:P}
    \resvec{P}{W}^i &= \begin{bmatrix} x & y & z \end{bmatrix}^T + \resvec{\rho}{W}^i\\
    \bm{v}^i &= \bm{v} + \bm{\omega} \times \bm{\rho}^i
\end{align}
where
\begin{subequations}
\begin{align}
    \resvec{\rho}{B}^{fl} &= \begin{bmatrix} \phantom{-}L_f & \phantom{-}\frac{e}{2} & -(h+R) \end{bmatrix}^T\\
    \resvec{\rho}{B}^{fr} &= \begin{bmatrix} \phantom{-}L_f & -\frac{e}{2} & -(h+R) \end{bmatrix}^T\\
    \resvec{\rho}{B}^{rl} &= \begin{bmatrix} -L_r & \phantom{-}\frac{e}{2} & -(h+R) \end{bmatrix}^T\\
    \resvec{\rho}{B}^{rr} &= \begin{bmatrix} -L_r & -\frac{e}{2} & -(h+R) \end{bmatrix}^T
\end{align}
\end{subequations}

Then, the relative extension and extension rate are computed as a function of the terrain height and slope at the previously determined position:
\begin{align}
    \label{eq:models:6dof:chi}
    \chi_i &= \frac
        {\resveccomp{P^i}{W}{z} - f(\resveccomp{P^i}{W}{x},\ \resveccomp{P^i}{W}{y})}
        {\resveccomp{\bar{\tau}^i}{B}{z}} \\
    \dot{\chi}_i &= \frac
        {\bar{\bm{\tau}}^i \cdot(\bm{v}^i - \chi_i \bm{\omega} \times \mathbf{b}_z)}
        {\resveccomp{\bar{\tau}^i}{B}{z}}
\end{align}
where $\bar{\bm{\tau}}^i$ is parallel to the normal vector, found by evaluating the slope of the terrain at the nominal contact patch position:
\begin{equation}
_{\mathcal{\scriptscriptstyle W}} \bar{\bm{\tau}}^i
\triangleq
\begin{bmatrix}
    f_x(\resveccomp{P^i}{W}{x},\ \resveccomp{P^i}{W}{y}) & f_y(\resveccomp{P^i}{W}{x},\ \resveccomp{P^i}{W}{y}) & 1
\end{bmatrix}^T
\end{equation}

Equation \eqref{eq:models:6dof:chi} determines $\chi_i$ such that $\resvec{P}{W}^i - \chi_i \mathbf{b}_z$ is coincident with a planar approximation of the terrain
constructed below the nominal contact patch position. This operation determines the length-scale mentioned in assumption A\ref{app:models:6dof:assumptions:planar},
and it is less strict than its extended single-track model counterpart.

With the extension and extension rate known, the suspension forces are calculated via
\begin{align}
    \label{eq:models:6dof:Fs}
    F_{s,i} &= \frac{g}{2}K_{zz,i} \\
    F_{k,i} &= \max(F_{s,i} - k_i \chi_i,\ 0) \\
    F_{b,i} &= \begin{cases}
        \max(-b_i \dot{\chi}_i,\ -F_{k,i}), & F_{k,i} > 0\\
        0, & F_{k,i} = 0
    \end{cases} \\
    F_{z,i} &= F_{k,i} + F_{b,i}
\end{align}
where
$F_{s,i}$ is the nominal tire loading,
$K_{zz,i}$ is given by either \eqref{eq:models:3dof:Kzzf} or \eqref{eq:models:3dof:Kzzr} depending on $i$,
$F_{k,i}$ is the additional force by an assumed linear spring,
$F_{b,i}$ is the additional force by an assumed linear damper,
and $F_{z,i}$ is the overall loading on tire $i$, acting in the $\mathbf{b}_z$ direction.
The springs are parameterized by coefficients $k_f$ and $k_r$ for the front and rear, respectively,
and the dampers are parameterized by coefficients $b_f$ and $b_r$ for the front and rear, respectively.

The suspension equations ensure that forces are only generated when the tire is in contact with the ground and that the overall force is never attractive.
This allows the model to operate without the assumption of persistent tire contact, i.e., the equations are still valid after tire liftoff.

The slip angle for tire $i$ is computed as
\begin{equation}
    \alpha_i = \arctan \left( \frac{\resveccomp{v^i}{B}{y}}{\resveccomp{v^i}{B}{x}} \right) - \delta_i
\end{equation}
where $\delta_i$ is $\delta$ for the front tires and zero otherwise.
With this and the vertical force known, the lateral force, $F_{y,i}$, may be computed using the tire model discussed in Sec. \ref{app:models:3dof}.

The summed forces required by \eqref{eq:models:6dof:A} are computed as
\begin{align}
    F_y &= \sum_i F_{y,i} \cos{\delta_i} \\
    F_z &= \sum_i F_{z,i}
\end{align}

The summed moments required by \eqref{eq:models:6dof:A} are computed as
\begin{align}
    \begin{bmatrix} M_x \\ M_y \\ M_z \end{bmatrix} = \sum_i \left(
        \resvec{\rho}{B}^i \times \begin{bmatrix} F_{x,i} \\ F_{y,i} \cos{\delta_i} \\ F_{z,i} \end{bmatrix} \right)
\end{align}
where $F_{x,i}$ is the constraining force required to maintain assumption A\ref{app:models:6dof:assumptions:longvel}, computed via
\begin{equation}
    F_{x,i} = \begin{cases}
        0, & i \in \{fl,\ fr\} \\
        \frac{M}{2}(\resveccomp{\dot{v}}{B}{x} -
            \resveccomp{g}{B}{x} +
            \resveccomp{\omega}{B}{y} \resveccomp{v}{B}{z} -
            \resveccomp{\omega}{B}{z} \resveccomp{v}{B}{y}), & i \in \{rl,\ rr\}
    \end{cases}
\end{equation}

\subsection{Energy Stability Margin Constraint} \label{sec:formulation:esm}
When using the single rigid body model, an additional limitation of the tire-liftoff constraint becomes apparent.
When the vehicle dynamics model is of a sufficient fidelity to predict tire liftoff from causes other than in-plane accelerations, the constraint is overconservative.
For example, the many small tire liftoffs that may occur when driving over terrain with high-frequency ruts do not pose a danger to the vehicle (see Fig. \ref{fig:liftoff}), but they are deemed unsafe by a tire-liftoff constraint.
This is not an issue for the \sota approach, since the extended single-track model does not predict these liftoffs, but it presents an issue for the proposed approach.

\begin{figure}
    \centering
    \includegraphics[width=.2\textwidth]{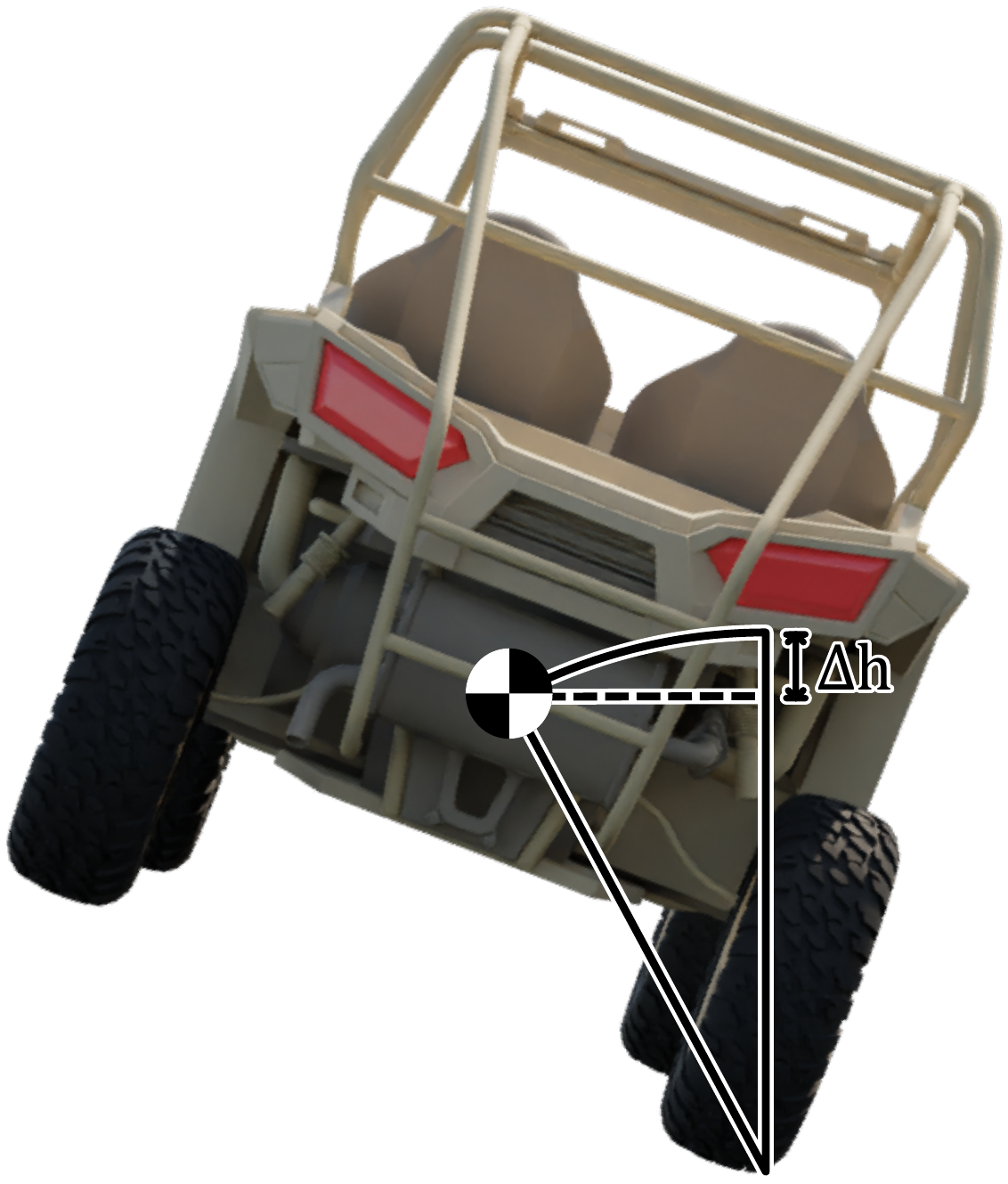}
    \caption{Relative height used to compute the Energy Stability Margin (ESM)}
    \label{fig:esm}
\end{figure}

Therefore, a constraint leveraging the energy stability margin (ESM) from \cite{messuriAutomaticBodyRegulation1985} is presented.
Fig. \ref{fig:esm} illustrates the application of ESM to a wheeled vehicle:
    it measures the minimum required potential energy change to bring the system to rollover or pitch-over,
    produced by rotating the vehicle about a line of contact with the ground such that the center of mass (CoM) lies directly over it.
With $\Delta h$ denoting the height difference of the CoM between the current and nearest unstable equilibrium state, ESM is computed via
\begin{equation}
    \label{eq:esm}
    U_\mathrm{ESM} \triangleq \begin{cases}
\phantom{-}M g \Delta h & \text{CoM is inside the vehicle footprint}\\
          -M g \Delta h & \text{CoM is outside the vehicle footprint}
\end{cases}
\end{equation}

ESM is at a maximum when the vehicle is sitting evenly on the ground, and it decreases to zero when the vehicle is perfectly balanced over an adjacent pair of wheels.
Additionally, the metric does not saturate; it can continue to go negative as rollover or pitch-over continues.
The ESM safety constraint simply asserts that the metric remains positive:
\begin{equation}
    U_\mathrm{ESM} > 0
    \label{eq:esm_ineq}
\end{equation}

The height difference, $\Delta h$, required to compute ESM via \eqref{eq:esm} is computed via the roll and pitch of the chassis.
The metric may be computed for the single rigid body model,
but it may not be applied to the extended single-track track model, since it lacks roll and pitch dynamics.

To simplify computation, it is assumed that rollover always requires less energy than pitch-over.
This is numerically true for the vehicle and terrain considered in this work, but it may not always hold.
It is also assumed that the location of the support polygon remains constant as the vehicle rolls and pitches, meaning this implementation of the metric neglects suspension deflection.
With these simplifications, $\Delta h$ is computed from 3-2-1 Euler angles via
\begin{equation}
    \label{eq:models:6dof:esm}
    \Delta h = \bar{R} \left( 1 - \sin\left(\lvert\phi\rvert+\bar{\phi}\right)\right) \cos(\theta)
\end{equation}
where
\begin{equation}
    \bar{R} \triangleq \sqrt{(h + R)^2 + \left(\frac{e}{2}\right)^2}
\end{equation}
and
\begin{equation}
    \bar{\phi} \triangleq \arctan\left(\frac{2(h+R)}{e}\right)
\end{equation}

\subsection{OCP Formulation} \label{sec:formulation:ocp}
At each iteration of the MPC outer loop, an OCP must be solved to determine the next control for the vehicle to execute.
Two problem formulations are created:
    one for a \sota baseline approach using the extended single-track vehicle model and a lateral-acceleration  constraint,
    and one for the proposed approach using the single rigid body vehicle model and an ESM constraint.
Both formulations each have an obstacle-avoidance constraint and
are identical outside of the vehicle dynamics model and rollover-prevention constraint.

A key component of the formulation is the cost function:
\begin{equation}
    J \triangleq \int_{t_0}^{t_f} L(\bm{\xi}(t), \bm{\zeta}(t)) dt + \Phi(\bm{\xi}(t_f))
    \label{eq:framework:cost}
\end{equation}
where the prediction horizon is given by $t_f - t_0$,
$L$ and $\Phi$ are the running and terminal cost functions,
and $\bm{\xi}$ and $\bm{\zeta}$ are the state and control vectors, respectively.
$\bm{\xi}$ is model-specific, given in \eqref{eq:models:3dof:state} and \eqref{eq:models:6dof:state} for the extended single-track and single rigid body models, respectively.
$\bm{\zeta}$ is identical between the models and given in \eqref{eq:models:3dof:control}.

The running cost includes two terms borrowed from \cite{yuNonlinearModelPredictive2021}:
\begin{equation}
    L_0(\bm{\zeta}) \triangleq w_t + w_c \dot{\delta}^2
    \label{eq:formulation:baseline_running_cost}
\end{equation} where
$w_t$ is the time-to-goal weight,
$w_c$ is the control weight,
and $\dot{\delta}$ is the steering rate.
The terminal cost is the weighted 2D Euclidean distance to the goal:
\begin{equation}
    \Phi(\bm{\xi}) \triangleq w_g \sqrt{(x - x_g)^2 + (y - y_g)^2}
\end{equation} where
$w_g$ is the distance-to-goal weight,
$x$ and $y$ describe the position of the vehicle's CoM,
and $x_g$ and $y_g$ describe the position of the goal.

The baseline running cost, $L_0$, minimizes time to goal and steering actuator usage, and the terminal cost penalizes terminal distance to the goal.
In this study
    $w_t$ is 5 \si{\costs\per\s},
    $w_c$ is to 8 \si{\costs\per\s} per \si{(\radian\per\s)\squared},
    and $w_g$ is 15 \si{\costs\per\m}.
These values were selected by manual tuning and observing the vehicle's navigation behavior on flat, level ground.

The full running cost, $L$, incorporates an additional term, $L_\mathrm{soft}$,
    which is the sum of several soft constraint cost rates and enables the implementation of safety constraints:
\begin{equation}
    L \triangleq L_0 + L_\mathrm{soft}
\end{equation}

\begin{figure}
    \centering
    \includegraphics[width=0.9\linewidth]{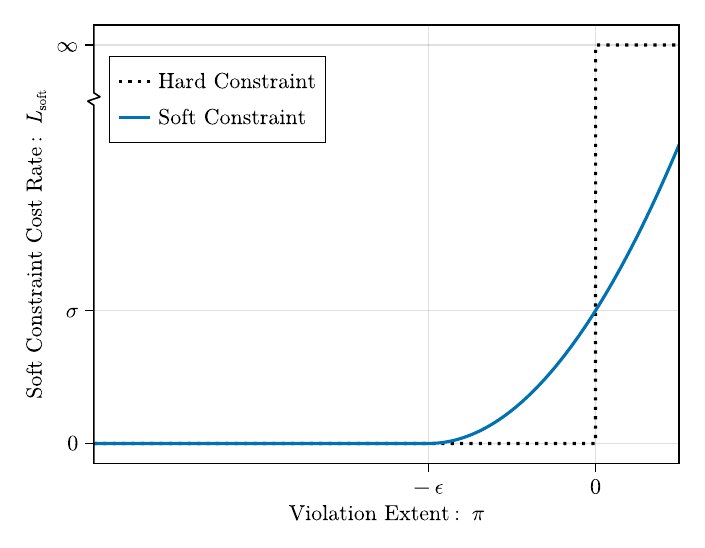}
    \caption{Normalized soft constraint formulation:
    The additional cost rate added by incremental violation of a soft constraint is shown
    and compared to the equivalent additional cost of violating a hard constraint.}
    \label{fig:framework:constraint1}
\end{figure}

Fig. \ref{fig:framework:constraint1} illustrates the normalized soft constraint formulation used in this work in comparison with a hard constraint approach.
Hard constraints are boolean functions, where all violations are equivalent.
In contrast, soft constraints provide a differential between violations of different magnitudes, promoting minimal constraint violation if a violation is inevitable.

Several types of soft constraints are applied, and they are expressed as normalized quadratic costs:
\begin{equation}
    L_\mathrm{soft} \triangleq \sum_{c \in \mathcal{C}} \sigma_c \max \left( 0, 1 + \frac{\pi_c}{\epsilon_c} \right)^2
\end{equation}
where $\mathcal{C}$ is the set of all applied constraints.
For constraint $c \in \mathcal{C}$,
$\pi_c$ measures the extent of violation, and $\epsilon_c$ and $\sigma_c$ are normalizing factors,
    shown in Fig. \ref{fig:framework:constraint1} for a single constraint.

The lateral-acceleration constraint is applied to the extended single-track model, and the ESM constraint is applied to the single rigid body model.
Both formulations also have a common set of distance-based constraints that prevent collision with obstacles and path boundaries.
The set of applied constraints for each formulation is therefore
\begin{align}
    \mathcal{C}_\mathrm{EST} &\triangleq \mathcal{C}_\mathrm{dist} \cup \{c_\resveccomp{a}{B}{y}\} \\
    \mathcal{C}_\mathrm{SRB} &\triangleq \mathcal{C}_\mathrm{dist} \cup \{c_\mathrm{ESM}\}
\end{align}
where subscripts EST and SRB stand for the extended single-track and single rigid body models, respectively.
$\mathcal{C}_\mathrm{dist}$ is the set of collision-avoidance constraints,
$c_\mathrm{\resveccomp{a}{B}{y}}$ is the lateral-acceleration constraint, and
$c_\mathrm{\mathrm{ESM}}$ is the ESM constraint.

Letting $\mathcal{P}$ denote the set of obstacle and path boundary perimeters
and $c_{\mathrm{dist}_{i,p}}$ denote a distance constraint between wheel $i$ and geometric feature $p$,
\begin{equation}
    \mathcal{C}_\mathrm{dist} \triangleq \{ c_{\mathrm{dist}_{i,p}} : i,p \in \{fl,\ fr,\ rl,\ rr\} \times \mathcal{P} \}
\end{equation}
Thus, each geometric feature generates a constraint for each wheel of the vehicle
This is done to approximate a constraint placed on the entire footprint of the vehicle.

The constraints, in their nominal, hard-constraint form, are
\begin{align}
        c_{\mathrm{dist}_{i,p}} &\triangleq \mathrm{dist}_{i,p} > 0 \\
    \label{eq:formulation:aby_ineq}
        c_\resveccomp{a}{B}{y} &\triangleq \lvert\resveccomp{a}{B}{y} - \resveccomp{g}{B}{y}\rvert < \resveccomp{\bar{a}}{B}{y} \\
    \label{eq:formulation:esm_ineq}
        c_\mathrm{ESM} &\triangleq U_\mathrm{ESM} > 0
\end{align}
where $\mathrm{dist}_{i,p}$ is the minimum 2D Euclidean distance between wheel $i$ and perimeter $p$,
the right-hand side of \eqref{eq:formulation:aby_ineq} is a simplified form of \eqref{eq:modeling-safety:aby_ineq},
and the right-hand side of \eqref{eq:formulation:esm_ineq} is identical to \eqref{eq:esm_ineq}.
To implement the above as soft constraints, a measure of violation is defined for each:
\begin{align}
        \label{eq:formulation:sdist}
    \pi_{\mathrm{dist}_{i,p}} &= -\mathrm{sdist}_{i,p} \\
    \pi_\resveccomp{a}{B}{y} &= \lvert\resveccomp{a}{B}{y} - \resveccomp{g}{B}{y}\rvert - \resveccomp{\bar{a}}{B}{y} \\
    \pi_\mathrm{ESM} &= -U_\mathrm{ESM}
\end{align}
where $\mathrm{sdist}_{i,p}$ is the signed minimum 2D Euclidean distance between wheel $i$ and perimeter $p$.
For an obstacle, a signed distance is negative when the test point falls inside the perimeter and is positive otherwise.
For a path boundary, the reverse is true.
The perimeters may be any arbitrary geometry so long as the signed distance can be computed for any test point.

\begin{figure}
    \centering
    \includegraphics[width=0.9\linewidth]{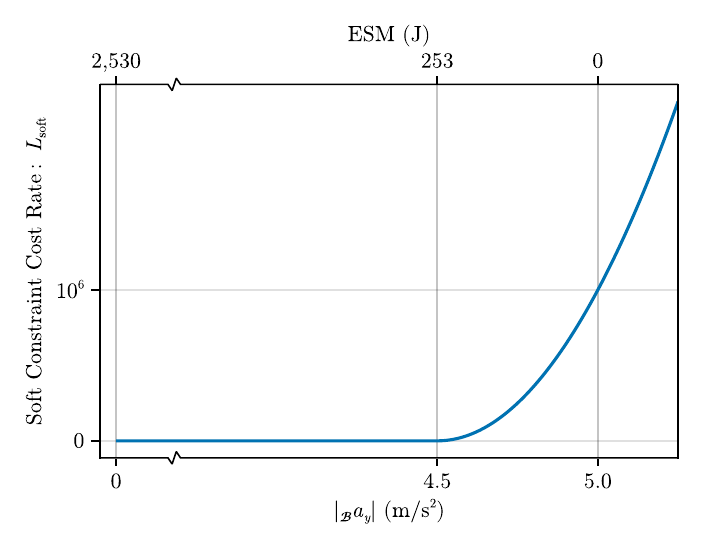}
    \caption{Safety constraint implementations:
    The rollover prevention constraints are normalized equivalently to minimize bias.
    Each constraint begins contributing cost at 10\% safety factor and has an identical value at full violation.
    The ESM constraint is shown with the top axis, and the lateral-acceleration constraint is shown on the bottom one.}
    \label{fig:framework:constraint2}
\end{figure}

For the rollover constraints, lateral-acceleration and ESM, the normalizing factors, $\epsilon_\resveccomp{a}{B}{y}$ and $\epsilon_\mathrm{ESM}$, are set such that each soft constraint begins contributing cost at a 10\% safety factor.
This is shown in Fig. \ref{fig:framework:constraint2}, and aims to minimize bias due to the disparate constraint types between the two formulations.

For both formulations, the obstacle and path boundary constraints use a normalizing factor, $\epsilon_{\mathrm{dist}_{i,p}}$, of \qty{0.25}{\m}.
This means they begin contributing cost when a wheel enters the region formed by inflating all perimeters by the same radius.

For all formulations and soft constraint types, the cost at exact violation, $\sigma_c$, is set to $10^6$ \si{\costs\per\s}.
A large value is selected so that avoiding a constraint will always take precedence over other costs, such as actuator usage.
Additionally, this means that obstacle and path boundary constraints are treated as being equally important to rollover constraints.

The final constraint considered is the maximum element-wise magnitude of the control input, $\bm{\zeta}$.
The first term, $\dot{\delta}$, is constrained by the actuation limits of the vehicle to a maximum magnitude of \qty{1}{\radian\per\s}.
The second term, $\resveccomp{\dot{v}}{B}{x}$, is constrained by the experimental design (discussed in Sec. \ref{sec:setup}) to always be zero.
Rather than treating these as constraints in the formulation, they can be satisfied by configuring the optimizer to only consider inputs from the set
    $\{\bm{\zeta} : \lvert\dot{\delta}\rvert \leq \qty{1}{\radian\per\s} ,\, \resveccomp{\dot{v}}{B}{x} = \qty{0}{\m\per\s\squared} \}$.

The navigation task is completed when the vehicle reaches the goal, which is specified as a circle in the horizontal plane;
i.e., the task is complete when the following is true:
\begin{equation}
    \sqrt{(x-x_g)^2 + (y-y_g)^2} \leq r_g
    \label{eq:formulation:goal}
\end{equation}
where $r_g$ is the goal radius, \qty{2.5}{\m} in this study.

The planning horizon is broken into $n_t$ segments of length $\Delta t$,
and the control inputs for the planning horizon are parameterized by a sequence of length $n_t$:
\begin{equation}
    [\bm{\zeta}]_{i=1}^{n_t} \triangleq [\bm{\zeta}_1, \bm{\zeta}_2, ..., \bm{\zeta}_{n_t-1}, \bm{\zeta}_{n_t}]
\end{equation}
The control inputs are fed to the plant via a zero-order hold (ZOH), where each $\bm{\zeta}_i$ is held for a duration of $\Delta t$, producing a step function.
$\Delta t$ is distinct from the internal period of any numerical integration schemes and from the period of the outer MPC loop.
In this study, $n_t$ is 16, and $\Delta t$ is \qty{0.25}{\s}, resulting in a planning horizon of \qty{4}{\s}.
A longer time horizon or shorter $\Delta t$ may result in better planning, but it increases the computational complexity of solving the OCP.

The solution to the overall OCP formulation is given by
\begin{equation}
    [\bm{\zeta}^*]_{i=1}^{n_t}
    \triangleq
    \argmin_{[\bm{\zeta}]_{i=1}^{n_t}}
    J([\bm{\zeta}]_{i=1}^{n_t})
    \quad\mathrm{s.t.}\quad
    \begin{aligned}
        &\dot{\bm{\xi}}(t) = f(\bm{\xi}(t), \bm{\zeta}(t)) \\
        &\bm{\xi}(t_0) = \bm{\xi}_0
    \end{aligned}
\end{equation} where
$f$ is the model-specific vehicle dynamics function and
$\bm{\xi}_0$ is the most recent vehicle state estimation.
Additionally, when integrating cost as defined in \eqref{eq:framework:cost},
$t_f$ is the lesser of $t_0 + n_t \, \Delta t$ or the minimum time such that \eqref{eq:formulation:goal} is satisfied.

\subsection{Real-Time Implementation}\label{sec:formulation:cuda}

The MPC implementation uses a sampling method to find an approximate solution to each OCP.
Each solution is found by drawing $N$ i.i.d. samples from the set of valid control actions, then independently computing the resulting cost, given in \eqref{eq:framework:cost}, for each sample.
The sample with the minimum cost is found and returned.
This operation is denoted an ``empirical argmin,''
and as $N$ increases, the empirical solution becomes a closer approximation of the true optimal solution.

Notably, this approach bears many similarities to MPPI \cite{williamsAggressiveDrivingModel2016},
which also uses random samples to approximate solutions to OCPs.
One key difference is that MPPI generates its solution as a weighted average of multiple samples,
and the empirical-argmin optimizer uses only a single sample.
In effect, the latter's approach is mathematically similar to
setting $\lambda$, a configuration parameter of MPPI, to zero.
Another is that MPPI is typically implemented with warm-starting,
but this work uses a uniform distribution for each iteration.

These differences can be beneficial in sample-constrained settings, but they are not necessary for the current work.
The GPGPU implementation of the proposed method is sufficiently computationally efficient
to enable real-time closed-loop operation of the empirical-argmin optimizer.

Compared to CPUs,
    GPUs have a much larger number of cores,
    each core has multiple resident threads that it can instantly switch between,
    and there is specialized texture memory hardware to perform bilinear interpolation and spatial caching \cite{NVIDIAAdaGPU}.
The implementation of the proposed method is written using a homogeneous multi-threading model where
each thread independently simulates a vehicle trajectory and evaluates cost through numerical integration.

A key limiting factor for this application is the register usage.
Multiple threads may share time on a single core, but they do not share registers.
Meaning, the combined register usage of all threads resident to a core must not exceed the hardware-limited register count.

Given this limitation, the total number of threads that may be simultaneously resident on a given GPU is
\begin{equation}
    N_\mathrm{resident,max} =
    \floor{\frac{\text{registers per core}}{\text{registers per sample}}} \, \text{core count}
\end{equation}
assuming an even distribution of registers between cores and threads.
The total number of samples can be expressed in terms of this quantity: $N = k \, N_\mathrm{resident,max}$,
    and, in theory, maximum sampling bandwidth is achieved with an integer $k$.

A CUDA C++ implementation of this algorithm compiled with the CUDA 12.8.0 toolkit requires 38 and 72 registers per sample for the extended single-track and single rigid body models, respectively.
Recent NVIDIA GPU architectures have 512 32-bit registers available to each core under a homogeneous workload \cite{NVIDIAAdaGPU},
    so $N_\mathrm{resident,max}$ is 7 times the number of cores for the single rigid body model.

The integration routines on the GPU are performed using a forward Euler method with a timestep of \qty{5}{\ms}.
Comparisons performed with various timesteps and an alterative \nth{4}-order Runge-Kutta method revealed
this method and timestep to offer the best real-time performance while maintaining accuracy for the single rigid body model.

Notably, the extended single-track model requires fewer registers than the single rigid body model and may be capable of a coarser timestep.
The former is simply caused by the computations being less complex, and the latter is caused by a lack of suspension modeling in the dynamics.
Therefore, the extended single-track model may be capable of larger sample sizes for the same runtime, but this study is performed with an identical number of samples and timestep for each formulation.
This is done to ensure similar convergence properties and to minimize differences between the formulations.

To enable efficient computation on the GPU, the $\mathrm{sdist}$ function specified in \eqref{eq:formulation:sdist} is pre-computed and summed for all features.
This is saved to a costmap that is interpolated at runtime, greatly improving efficiency.
Additionally, the randomly sampled controls are directly generated on the GPU via the cuRAND library, and the $\argmin$ operation is also performed on the GPU through a parallelized algorithm.

Real-time suitability of the proposed formulation is evaluated using Dell Precision 5690 laptop with an Intel Core Ultra 9 185H CPU, 64 GiB RAM, and an NVIDIA RTX 5000 Ada Mobile GPU.
This GPU has 9728 cores, making $N_\mathrm{resident,max}$ \num{68 096}. $k$ is set to 3, making $N$, the total number of samples per OCP, \num{204 288}.
Under this configuration, the CUDA C++ implementation achieves a maximum execution rate of \qty{80}{\Hz} with a corresponding total system power draw of \qty{140}{\W},
while using the single rigid body model.

Equivalently stated, the implementation is capable of evaluating over 11 million trajectory samples per second on this hardware.
It is possible to increase or decrease the update rate by varying the number of samples per OCP.
The specific value of $k$ used in these experiments was selected to maximize convergence for challenging OCPs by providing an overestimate of the required number of samples.
The sampling rate is also dependent on the number of cores available on the GPU.
For example, this formulation achieves approximately 21 million trajectory samples per second on the NVIDIA RTX 6000 Ada Generation desktop GPU with \num{18 176} cores.

\section{Setup of Navigation Trials}\label{sec:setup}
Multiple trials of several closed-loop navigation tasks are used to evaluate the relative performance of the proposed formulation compared to the \sota approach.
The tasks require the traversal of rugged off-road terrain, at speeds ranging from 4 to 10 \si{\m\per\s}.
To assess how well rollover and obstacle collision are avoided as a function of speed, the prescribed longitudinal velocity is fixed for each trial, making it an independent variable in the analysis.
This is achieved by using a PID controller to maintain speed via throttle actuation.
The MPC formulation is used to control steering only,
meaning all candidate controls are generated with the second term of \eqref{eq:models:3dof:control} set to zero.

Trials are performed both in simulation and with a full-size physical vehicle, and the simulations are configured to closely mimic the real-world testing conditions.
This enables both a larger number of trials to be performed and the trials to be carried out at higher speeds than is safely possible with the real vehicle.
A total of 9900 simulated trials are conducted at speeds ranging from 5 to 10 \si{\m\per\s}, and a total of 48 physical trials are conducted at speeds ranging from 4 to 5.5 \si{\m\per\s}

\begin{figure}
    \centering
    \includegraphics[width=0.9\linewidth, trim={0cm 6.5cm 9cm 0cm}, clip]{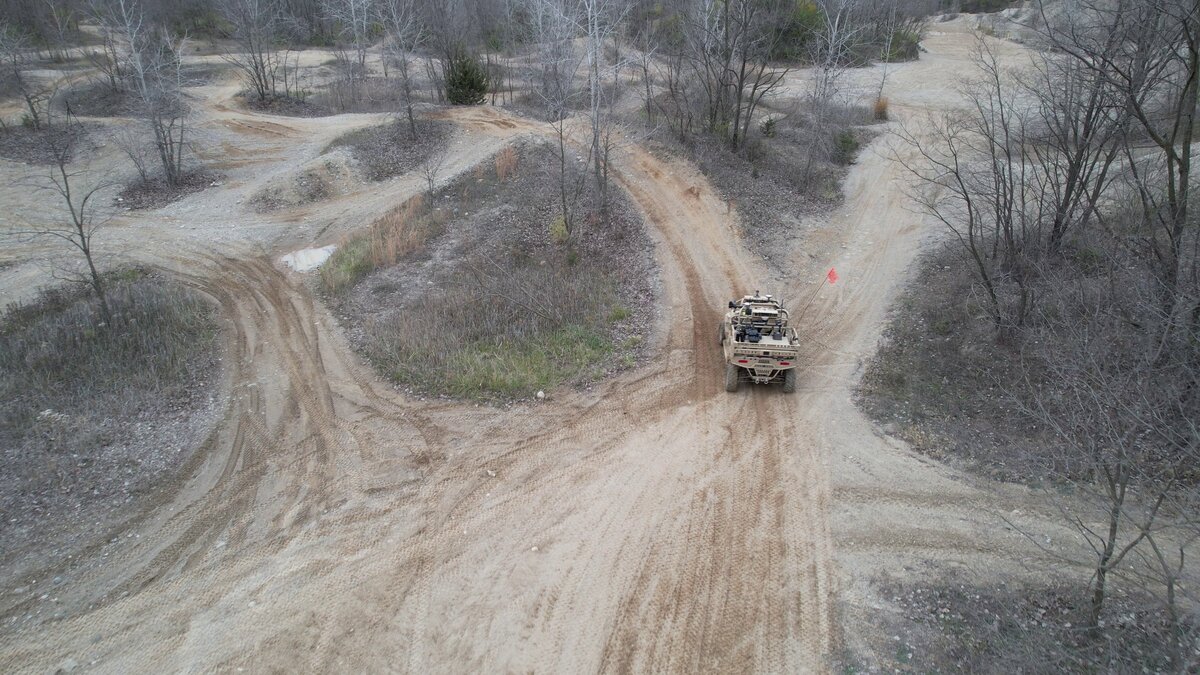}
    \caption{The physical location and vehicle used for navigation trials}
    \label{fig:setup:bundy}
\end{figure}

The navigation trials take place in a hilly and rugged area with many intersecting off-road trails.
This location, found at the Bundy Hill Off-Road Park in Jerome, Michigan, is shown in Fig. \ref{fig:setup:bundy} along with the experimental vehicle.

To perform the simulated trials, a 3D reconstruction of the physical terrain is created by MTRI Inc.
Images are captured by a UAV and processed with the Propellor photogrammetry software.
Obstacles and path boundaries are manually identified in the scanned data.

The trials include three scenarios, two of which are the same route, but run in opposite directions.
The simulated trials use all three scenarios equally, whereas the physical trials use one scenario.
The navigation scenarios and terrain are described and shown in Sec. \ref{sec:setup:task}.

The scanned terrain provides the trajectory planners with \textit{a priori} terrain maps, and it is also used by the plant co-simulator in the simulated trials.
Due to the assumptions made by each model, it is necessary to artificially smooth rough terrain before using it with the MPC formulation.
Further justification for this step, the procedure, and visualizations of the terrain representations are provided in Sec. \ref{sec:setup:smoothing}.

The trials are designed to emulate complete autonomous vehicle systems that are identical except for the MPC formulation.
The required supporting software and hardware tools and the plant vehicle are described in
Secs. \ref{sec:setup:vortex} and \ref{sec:setup:mrzr} for the simulated and physical trials, respectively.

\subsection{Scenario Descriptions}\label{sec:setup:task}
The three scenarios are constructed as follows:
\begin{enumerate}[\hspace{5pt}{Scenario} 1)]
    \item primary route, forward direction
    \item primary route, reverse direction
    \item secondary route, forward direction
\end{enumerate}
The secondary route is more challenging, and experiments running it in a reverse direction are not included since neither formulation was able to complete it at any speed without obstacle violation.

Each forward route begins with a climb up the hill pictured in Fig. \ref{fig:setup:bundy}, followed by traversal over a series of bumps as it follows the ridge also pictured in Fig. \ref{fig:setup:bundy}.
The two forward routes diverge at this point, with the primary route making an earlier, banked right turn before straightening out to reach the goal,
and the secondary route making a later, sharper right turn to follow a second, rougher path to the goal.

The reverse route follows an identical path to the primary forward route, but the location of the start and goal is switched.
Turns are made in opposite directions, and the route ends with a descent down the hill.

\begin{figure}
    \centering
    \includegraphics[width=\linewidth]{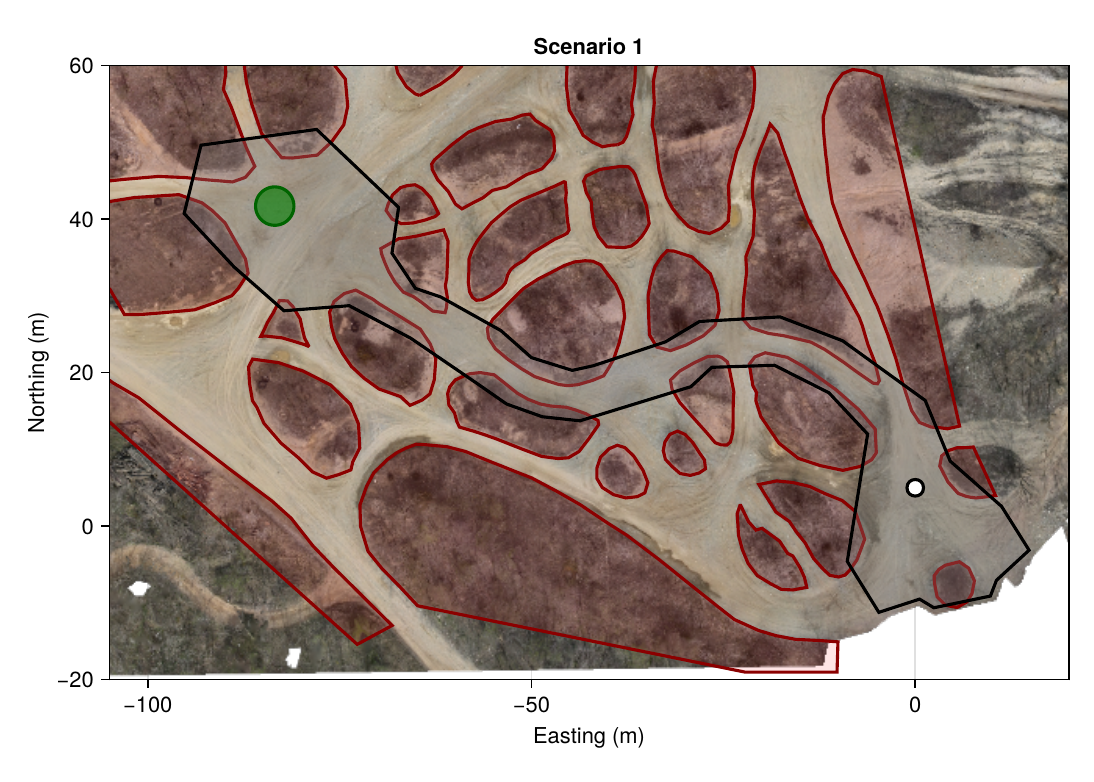}
    \\\vspace{-4pt}
    \includegraphics[width=\linewidth]{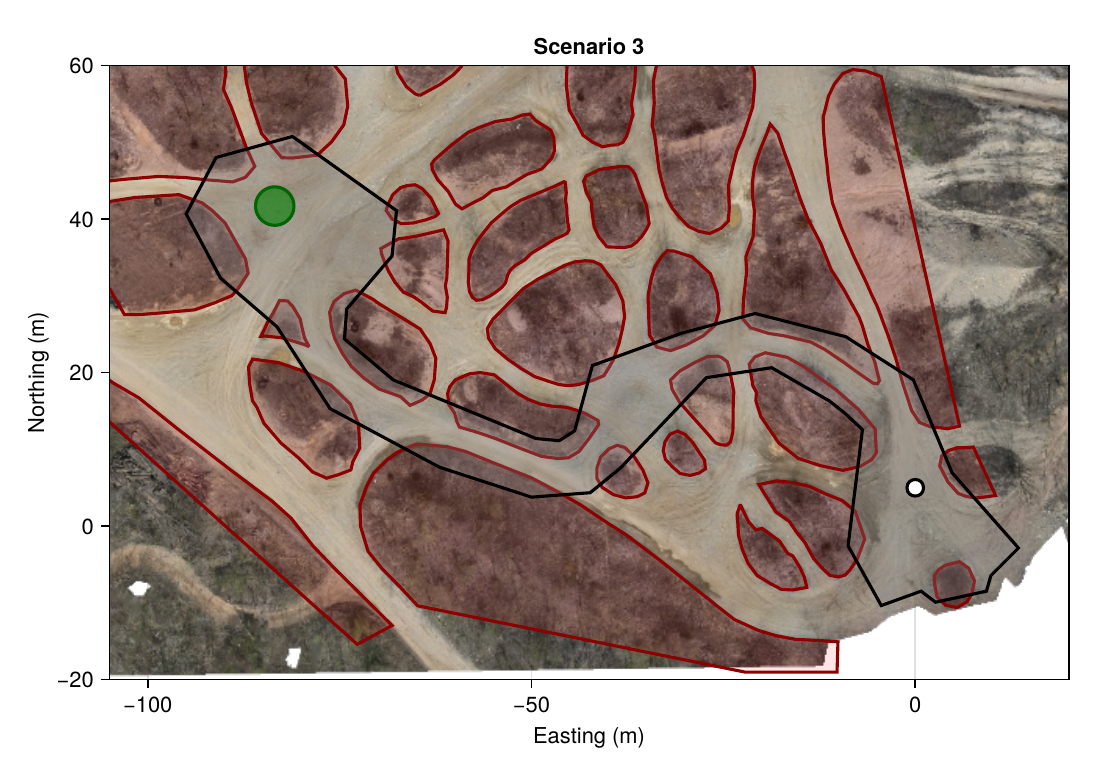}
    \caption{
        Route geometry for scenarios 1 and 3, visualized over reconstructed terrain imagery.
        The vehicle must navigate from the starting point (white circle) to the goal (green circle),
        while staying inside the path boundary (black polygon) and outside all obstacles (red polygons).
    }
    \label{fig:setup:scenarios}
\end{figure}

\begin{figure*}
    \centering
    \includegraphics[width=0.23\linewidth]{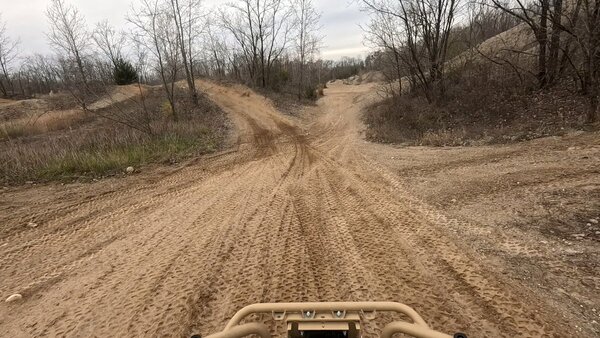}
    \includegraphics[width=0.23\linewidth]{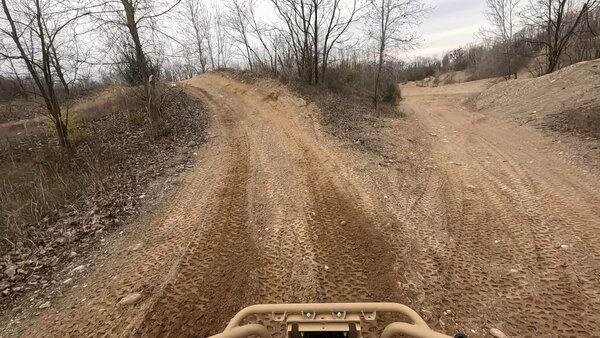}
    \includegraphics[width=0.23\linewidth]{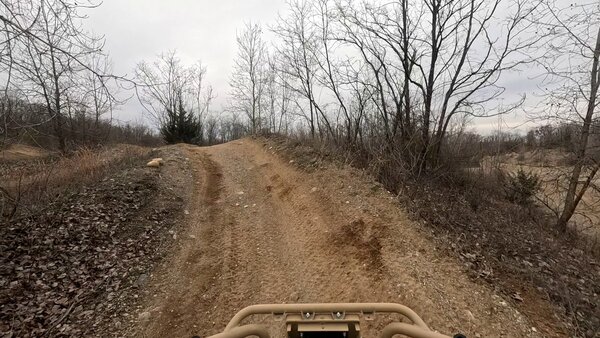}
    \includegraphics[width=0.23\linewidth]{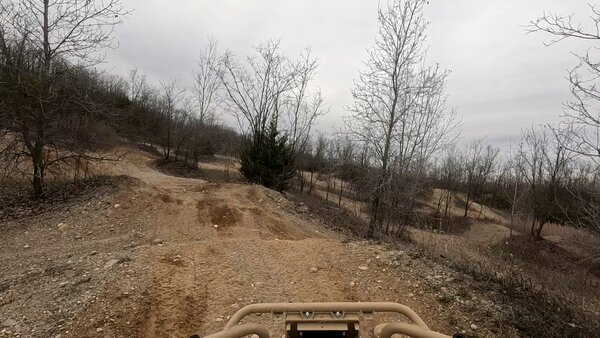}
    \\\vspace{4pt}
    \includegraphics[width=0.23\linewidth]{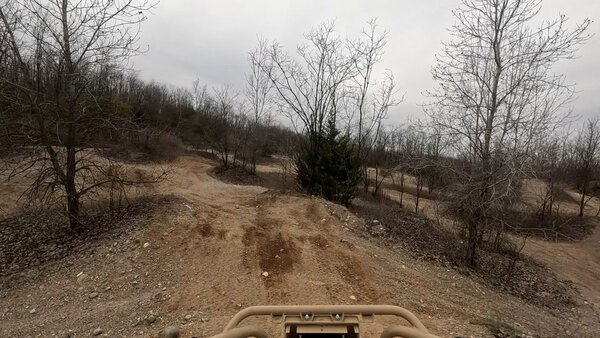}
    \includegraphics[width=0.23\linewidth]{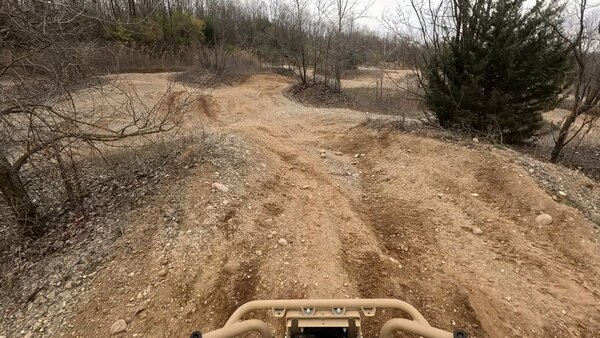}
    \includegraphics[width=0.23\linewidth]{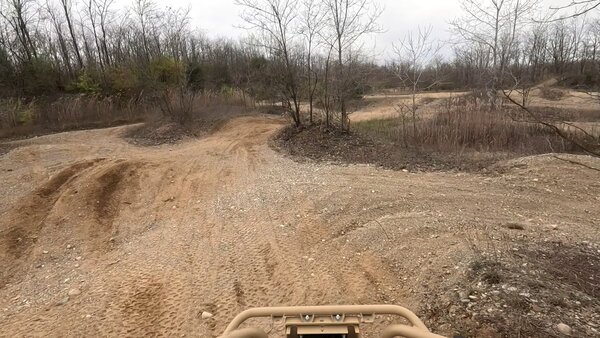}
    \includegraphics[width=0.23\linewidth]{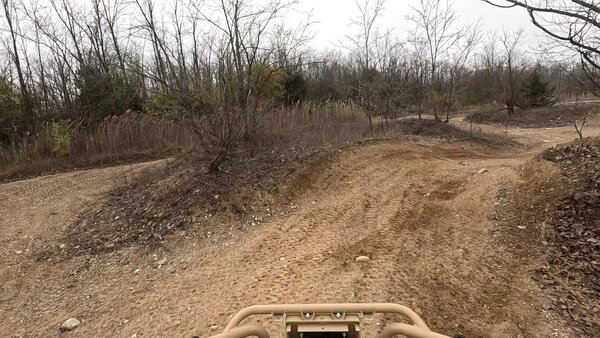}
    \\\vspace{4pt}
    \includegraphics[width=0.23\linewidth]{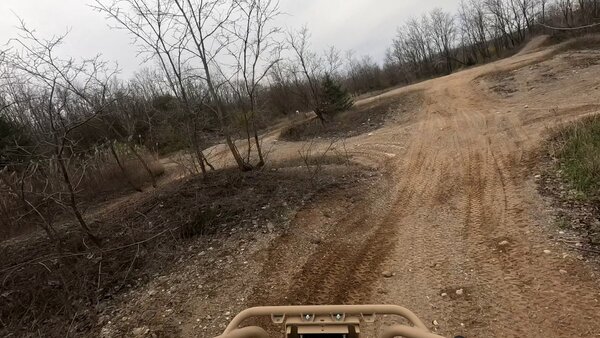}
    \includegraphics[width=0.23\linewidth]{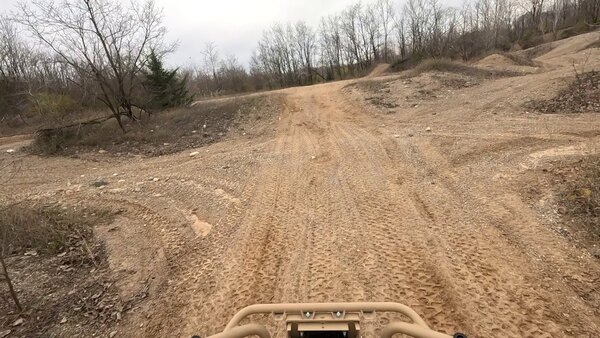}
    \includegraphics[width=0.23\linewidth]{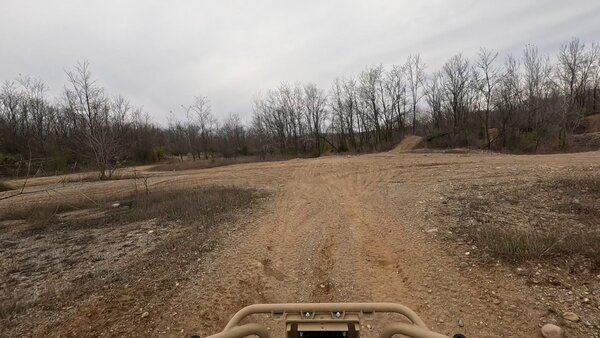}
    \includegraphics[width=0.23\linewidth]{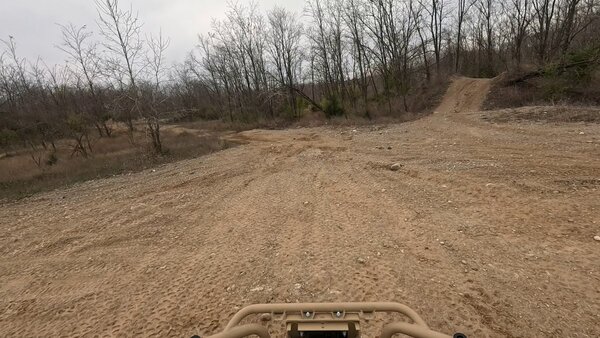}

    \caption{
        Snapshots of terrain encountered  while autonomously driving scenario 1 (viewed left-to-right, top-to-bottom).
        The images span approximately 25 seconds of footage from a physical trial of the proposed formulation carried out at a speed of 5 \si{\m\per\s}.
    }
    \label{fig:setup:task1}
\end{figure*}

Each scenario starts with the vehicle already in motion at the starting point (due to the prescribed constant longitudinal velocity), and lasts 20 to 30 seconds depending on speed.
The route geometries for scenarios 1 and 3 are visualized via bird's-eye views in Fig. \ref{fig:setup:scenarios}, and
Fig. \ref{fig:setup:task1} provides images of the physical terrain on scenario 1 as the vehicle navigates it.

\subsection{Terrain Smoothing}\label{sec:setup:smoothing}
It is incorrect to simulate the models on terrains that do not meet the level of smoothness assumed for the model.
If planning is required over terrain that is too rough for the model, and the terrain estimation captures this roughness, then it must be artificially smoothed before performing planning.

The reasoning for this is as follows:
    the extended single-track model considers the chassis to be constantly aligned with the terrain's normal and permanently offset from the terrain by a fixed distance along the normal.
This offset is performed relative to the terrain directly below the CoM.
If the terrain is not smooth relative to the length of the chassis, the model may improperly roll and pitch the chassis due to the local variation below the CoM.
The single rigid body model computes suspension deflection by projecting a vector from the wheel location onto a virtual plane located and linearized beneath the wheel.
Similarly, if the terrain is not smooth relative to the size of the wheel, the model may improperly compute suspension deflection, leading to incorrect suspension behavior.
Both models require the terrain to be smooth, albeit to different extents, such that a point measurement describes the average behavior over a certain length scale.

The reconstructed terrain captures fine details and is too rough for either model.
Therefore, a Gaussian smoothing kernel with a model-dependent standard deviation is used to artificially smooth the terrain.
Gaussian kernels act as low-pass filters, removing the incompatible high-frequency components of the terrain profile while leaving the compatible low-frequencies.
A \qty{1.5}{\m} standard deviation is used for the extended single-track model, since it assumes terrain to be planar at the length-scale of the vehicle; and
a \qty{0.3}{\m} standard deviation is used for the single rigid body model, since it assumes terrain to be planar at the length-scale of a wheel.

\begin{figure}
    \centering
    \includegraphics[width=0.32\linewidth, trim={5.3333cm 2.6666cm 8cm 1.3333cm}, clip]{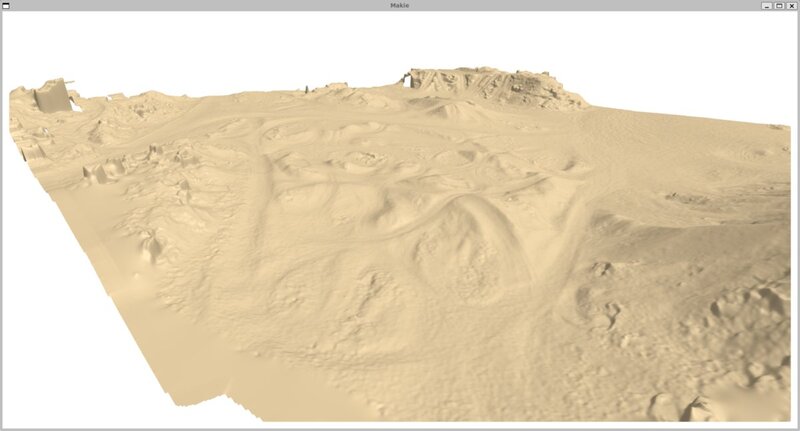}
    \includegraphics[width=0.32\linewidth, trim={5.3333cm 2.6666cm 8cm 1.3333cm}, clip]{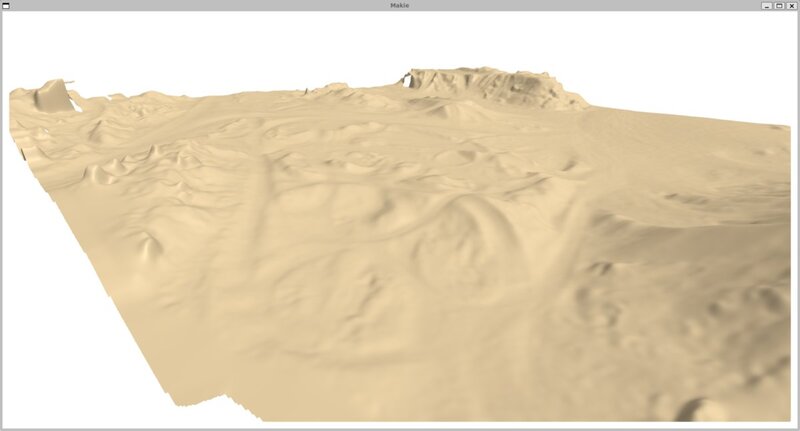}
    \includegraphics[width=0.32\linewidth, trim={5.3333cm 2.6666cm 8cm 1.3333cm}, clip]{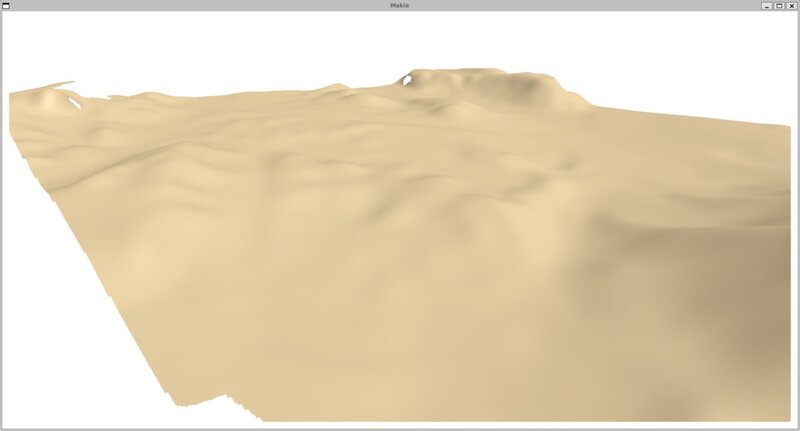} \\
    \caption{3D rendering of terrain at three levels-of-detail (LODs).
        Left: plant LOD, Center: single rigid body LOD, Right: extended single-track LOD.}
    \label{fig:setup:smoothing}
\end{figure}

This procedure generates two additional representations of the terrain geometry, which are referred to as levels-of-detail (LODs).
The ground-truth terrain representation, before any modifications are performed, is denoted as the plant LOD.
The model-specific terrains generated by the smoothing procedures are denoted as the single rigid body and extended single-track LODs.
Each of the terrain LODs are visualized in Fig. \ref{fig:setup:smoothing}.

\subsection{Plant Co-Simulator}\label{sec:setup:vortex}
The simulated trials are performed using CM Labs' Vortex Studio multi-body simulator \cite{VortexStudio}.
The tool provides a high-fidelity vehicle representation, including drivetrain, tire, and suspension modeling.
The plant vehicle model is created by Traxara Robotics as a digital replica of the full-scale vehicle used in physical tests (see Sec. \ref{sec:setup:mrzr}).
The simulated trials use an off-road tire model tuned by Traxara Robotics to match measurements from navigation on the fine-grained dry soil at Keweenaw Research Center in Calumet, MI.

The inputs to the simulated vehicle are steering, throttle, and brake actuator efforts.
The steering effort is produced by integrating and scaling the steering rate-of-change output from the MPC formulation,
and a PID controller maintains the longitudinal speed of the vehicle via throttle effort control.
The simulated brake actuator is not used.

The output from the simulated vehicle is a 3D odometry measurement of the chassis at a rate of \qty{100}{\Hz}.
The odometry measurement is used to form the vehicle state vector,
    which is provided to the planner as-is.
There is no simulated sensor noise or communication latency.

The planner runs at a fixed frequency of \qty{25}{\Hz}, significantly under the maximum real-time rate measured in Sec. \ref{sec:formulation:cuda}.
This adjustment is made to simplify synchronization with the \qty{100}{\Hz} plant simulator and to prevent full GPU utilization, leaving room for expansion via additional modules such as vision processing.
Simulation consists of a number of iterations, where at each iteration,
    a state measurement is provided to the planner from the plant,
    the planner computes a control action,
    and the plant simulates \qty{40}{\ms} of motion.
This procedure enables synchronized, repeatable simulations that may run faster than real-time, but it does not model the computational delay of the planner.

The simulation is configured such that collision with obstacles do not halt the simulation.
If the vehicle violates an obstacle constraint, the collision is noted, but the simulated trial proceeds without modification.
The collision physics of the obstacles are not simulated, and they have no effect on the vehicle dynamics.

The simulated vehicle states and control actions are logged during simulation, along with the final outcome of the trial.
The possible outcomes are
   (i)  success, meaning the vehicle reached the goal without experiencing rollover nor collision with obstacles or the path boundary;
   (ii) non-successful goal acquisition, meaning the vehicle reached the goal without experiencing rollover, but collision with obstacles or the path boundary occurred;
    (iii) rollover, meaning the chassis roll or pitch exceeded \ang{72}\footnote{The selected value of this threshold must be sufficiently high to ensure the vehicle has rolled over with no chance of recovery,
but it must also be low enough to trigger before the simulation is terminated for other reasons.
So long as both conditions are satisfied, the approach is robust to variation in this parameter,
as the roll angle of vehicles experiencing unrecoverable rollover will grow rapidly.}; and
    (iv) timeout, meaning \qty{60}{\s} of simulated time elapsed without rollover or reaching the goal.

\subsection{Experimental Off-Road Vehicle}\label{sec:setup:mrzr}
The physical experiments are performed using the full-scale Polaris MRZR D4 ultra-light off-road vehicle shown in Fig. \ref{fig:setup:bundy}.
The vehicle is configured for autonomous control of the throttle, brake, and steering actuators via a ROS1 Noetic interface.
Identical to the simulated trials, steering effort is controlled by the MPC formulation, and a PID controller maintains the longitudinal speed of the vehicle.

The 3D odometry measurements are produced at \qty{100}{\Hz} by an Oxford Technical Services RT3000v3 GNSS/INS.
The sensor is configured with dual antennas and uses NTRIP for RTK corrections.
During testing, the unit's self-reported accuracies (1 standard deviation) are \qty{3.6}{\cm} for position and \qty{2}{\cm\per\s} for velocity.
The GNSS/INS sensor is the only sensor used during the experiments; terrain and obstacle data is known \textit{a priori} from the 3D terrain reconstruction.

The local trajectory planner is run from a Dell Precision 5570 laptop with an Intel i7-12700H CPU, 64 GiB RAM, and an NVIDIA RTX A2000 Mobile GPU mounted in the passenger seat.
This system is several generations older than that used for the real-time feasibility studies in Sec. \ref{sec:formulation:cuda},
    so the extended single-track and single rigid body formulations have execution rates of \qty{10.42}{\Hz} and \qty{8.48}{\Hz}, respectively,
    when run at their maximum hardware-supported frequencies.
These speeds are still sufficient for real-time performance, but better integration of computer resources onto the experimental vehicle could reduce these discrepancies in the future.

Experiments are performed using an in-vehicle safety driver.
Once each trial begins, the only action available to the operators is to terminate the trial by taking full manual control.
It is not possible for operators to partially override or influence the trial without a full takeover.

The vehicle states and control actions are logged during trials, along with an outcome of either success or takeover.
Takeovers are reserved for extreme circumstances;
    if it is determined that manual control is taken prematurely, the trial is repeated or removed from the dataset.

\section{Results}\label{sec:results}
\subsection{Simulated Closed-Loop Trials}\label{sec:results:sim}

The simulated trials consider 9900 individual runs.
Each of the three scenarios are performed at 11 speeds evenly spaced between 5 and \qty{10}{\m\per\s}, yielding 33 combinations.
Each of these is simulated across three setups, producing 99 configurations.
Each configuration is simulated 50 times per formulation with randomly generated initial conditions.
The random initial conditions are matched to ensure
    the results reflect each formulation's general ability to complete the task,
    rather than being influenced by small changes in initial conditions that a given formulation may find favorable or unfavorable.

Since the single rigid body model requires less smoothing compared to the extended single-track model,
    its formulation receives a more detailed representation of rough terrain.
This may disadvantage the extended single-track formulation, since it is subjected to increased disturbance.
To prevent such biasing of the results, the terrain LOD provided to the planners and plant is varied across the three setups.

The three setups are described in Table \ref{tab:closed-loop:sim:setup}.
To summarize,
    setup 1 is the most realistic for replicating the physical scenario, but it results in the formulations being provided different information.
Setup 3 provides both formulations with identical information, eliminating that factor from the analysis, but it is the least realistic, as can be seen in Fig. \ref{fig:setup:smoothing}.
Setup 2 provides a middle ground between the two extremes.

\begin{table*}
    \centering
        \caption{Terrain levels of detail (LODs) used in each setup of the simulated study}
    \begin{tabular}{cccc}
        \hline
        Setup & Plant Co-Simulator & Single Rigid Body Formulation& Extended Single-Track Formulation \\ \hline
        1 & plant LOD & single rigid body LOD & extended single-track LOD \\
        2 & single rigid body LOD & single rigid body LOD & extended single-track LOD \\
        3 & extended single-track LOD & extended single-track LOD & extended single-track LOD
    \end{tabular}
    \label{tab:closed-loop:sim:setup}
\end{table*}

\begin{figure}
    \centering
    \includegraphics[width=\linewidth]{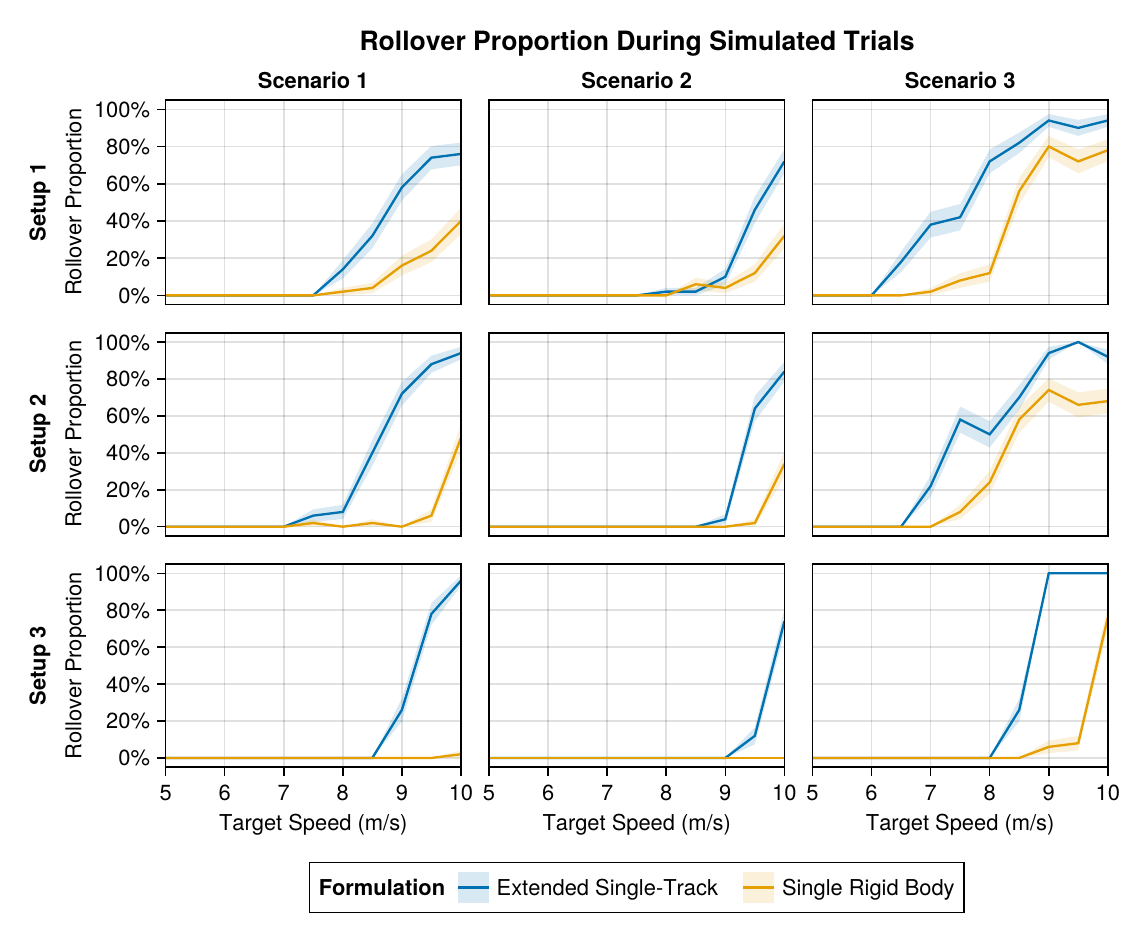}
    \caption{Results of simulated trials evaluating local trajectory planner performance.
        Rollover proportion is plotted as a function of speed, and each row provides results of a single setup, ranging from most realistic (top) to most equivalent (bottom).
        The single rigid body formulation consistently outperforms the extended single-track formulation in regions where a visible difference beyond the bounds of uncertainty exists,
            indicating that it has superior ability to ensure safety against rollover.}
    \label{fig:closed-loop:sim:rollover}
\end{figure}

Fig. \ref{fig:closed-loop:sim:rollover} presents the results of the simulated trials, plotting each formulation's proportion of trials resulting in rollover as a function of speed.
The three columns show the results from each of the three scenarios, and the three rows show the results from each of the three setups.
The sample mean is indicated with a solid line, and uncertainty bounds ($\pm$1 standard error) are indicated via the shaded regions.

It is evident in Fig. \ref{fig:closed-loop:sim:rollover} that in all areas where a visible difference beyond the bounds of uncertainty exists,
    the single rigid body formulation experiences the same or better performance compared to the extended single-track formulation.
That is, at a given speed, the proposed approach is found to meet or exceed the \sota baseline's ability to prevent rollover.
This difference is attributed to the proposed approach's capacity to model and mitigate a larger set of rollover types.
This result also shows that the additionally modeled rollover types are indeed relevant for high-speed off-road vehicles traveling on rugged non-planar terrain.

The first row, the most realistic setup, shows this result to be true for a high-fidelity simulation of an autonomous vehicle system on realistic rough off-road terrain.
The second and third rows, which reduce fidelity in order to eliminate discrepancies in comparative terrain knowledge,
    show that the result is not purely caused by the proposed approach's ability to use higher fidelity terrain representations.
The ability to use higher fidelity terrain representations may be a contributing factor to increased safety, but it is not the sole cause.

\begin{figure}
    \centering
    \includegraphics[width=\linewidth]{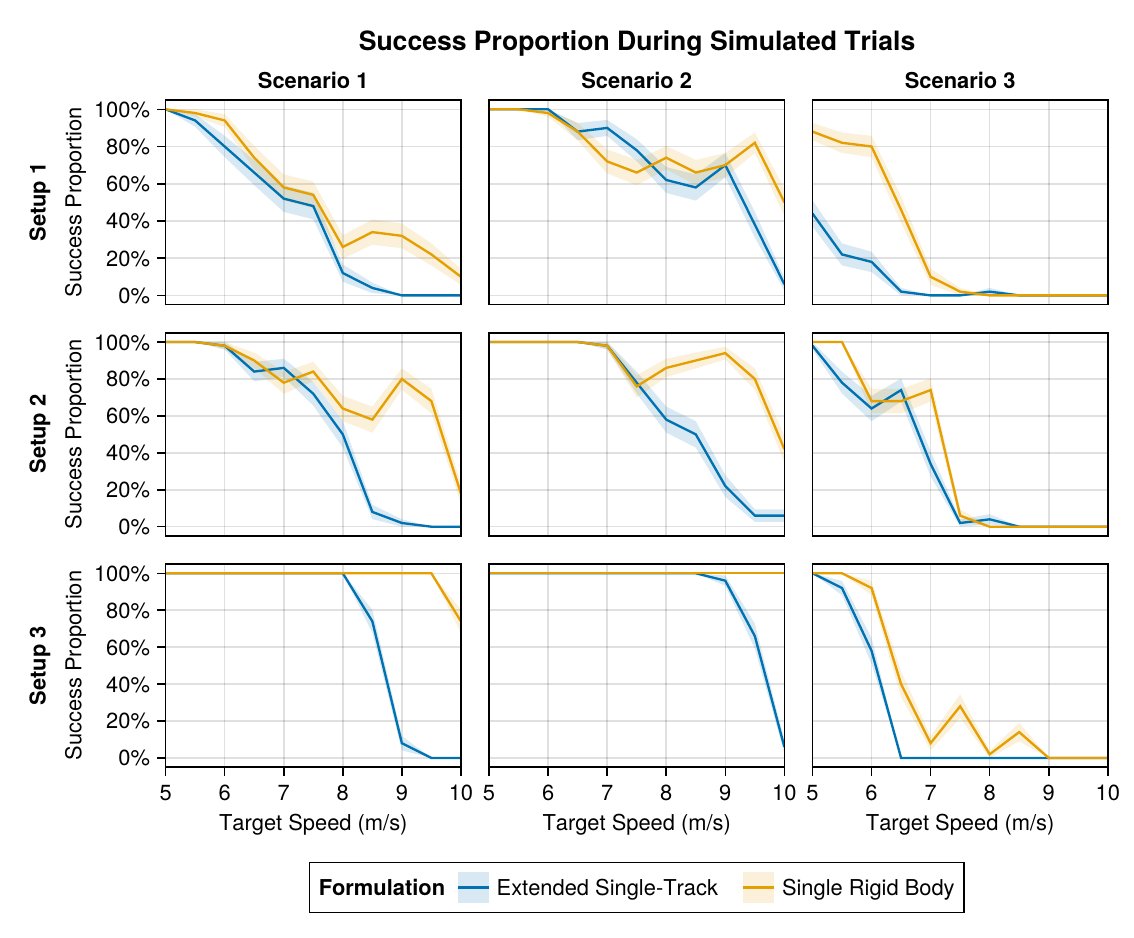}
    \caption{Results of simulated trials evaluating local trajectory planner performance,
        Success proportion is plotted as a function of speed, and each row provides results of a single setup, ranging from most realistic (top) to most equivalent (bottom).
        Superior performance is more often achieved by the single rigid body formulation than the extended single-track formulation,
            indicating that its superior ability to ensure safety against rollover does not come at the expense of success.}
    \label{fig:closed-loop:sim:success}
\end{figure}

Rollover proportion is the key metric of interest in this comparison, simply because the proposed approach primarily differs in its theoretical ability to prevent rollover.
However, success proportion is also important to consider, as it can sometimes appear in a trade-off with safety, such as discussed in Sec. \ref{sec:modeling-safety:discussion}.
Therefore, Fig. \ref{fig:closed-loop:sim:success} plots each formulation's proportion of trials resulting in success as a function of speed.

Fig. \ref{fig:closed-loop:sim:success} indicates that the single rigid body formulation has a higher success proportion than the extended single-track formulation more often than the reverse
    (52\% compared to 8\%, respectively; they are equal in the remaining 40\%).
A non-success can be caused by antecedents other than rollover, so the lack of absolute consistency is not a detractor.
These data indicate that the enhanced rollover prevention of the proposed formulation does not come at the expense of mobility.

Notably, both formulations tend to experience low success proportions at the top speeds of the trials.
This indicates that the scenarios are appropriately challenging to capture the upper limit of performance in extreme environments.
Interestingly, the plots decrease non-monotonically, meaning that increasing speed sometimes improves success proportion rather than worsening it.
This is not unexpected; the closed-loop behavior is complex due to the extreme terrain and nonlinear vehicle and suspension dynamics captured in the plant simulator.
For example, vibrations from traversing rough terrain may primarily manifest in the chassis at low speed then move to the wheels as the speed, and therefore excitation frequency, increases.
The decreased chassis vibration may make navigation easier, and this could be a potential cause of the non-monotonicity.

\subsection{Physical Closed-Loop Trials}\label{sec:results:real}
As opposed to the simulated trials, the physical trials consist of only scenario 1.
This decision is made for safety purposes, as the primary route is less challenging, and an autonomous hill ascent is deemed less dangerous than descent.

Notably, the empirical value of $\resveccomp{\bar{a}}{B}{y}$, the maximum lateral acceleration the vehicle may support without rollover, is \qty{5.0}{\m\per\s\squared} for the experimental vehicle.
Dividing by \qty{9.81}{\m\per\s\squared}, the right-hand side of \eqref{eq:modeling-safety:mu_ineq} is found to be \num{0.51}.
Additionally, $\mu$ is estimated as \num{0.4} at the testing location.
This is less than \num{0.51}, so as discussed in \ref{sec:modeling-safety:analysis:violation},
    the rollover-prevention constraint is permanently inactive for the baseline approach during the physical experiments.

The trials are run at speeds from 4 to \qty{5.5}{\m\per\s}, with the exact distribution shown in Table \ref{tab:closed-loop:real:trials}.
Due to time limitations in the testing area, the number of trials per speed vary.

\begin{table}
    \centering
    \caption{Distribution of trials during physical experiments}
       \begin{tabular}{ccc}
        \hline
        Target Speed & Trials per Formulation \\ \hline
        \qty{4.0}{\m\per\s} & 5 \\
        \qty{4.5}{\m\per\s} & 9 \\
        \qty{5.0}{\m\per\s} & 5 \\
        \qty{5.5}{\m\per\s} & 5 \\
    \end{tabular}
    \label{tab:closed-loop:real:trials}
\end{table}

\begin{figure}
    \centering
    \includegraphics[width=\linewidth]{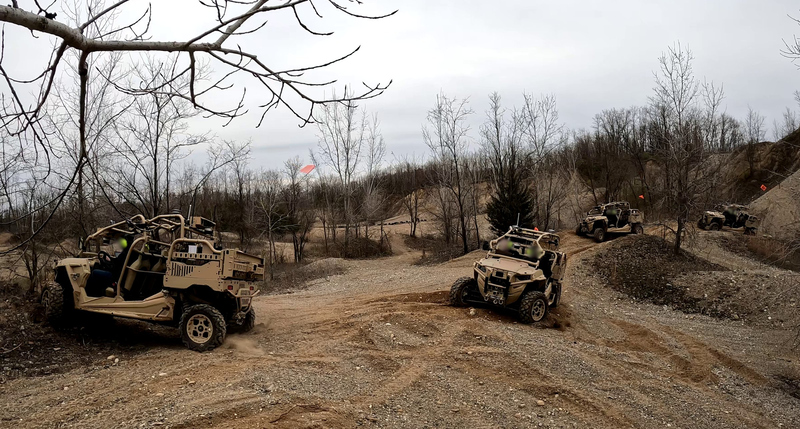}
    \caption{
        Composite time-lapse image of the failure experienced by the extended single-track formulation during a scenario 1 trial at \qty{5}{\m\per\s}}
    \label{fig:closed-loop:real:failure-composite}
\end{figure}

\begin{figure}
    \centering
    \includegraphics[width=\linewidth]{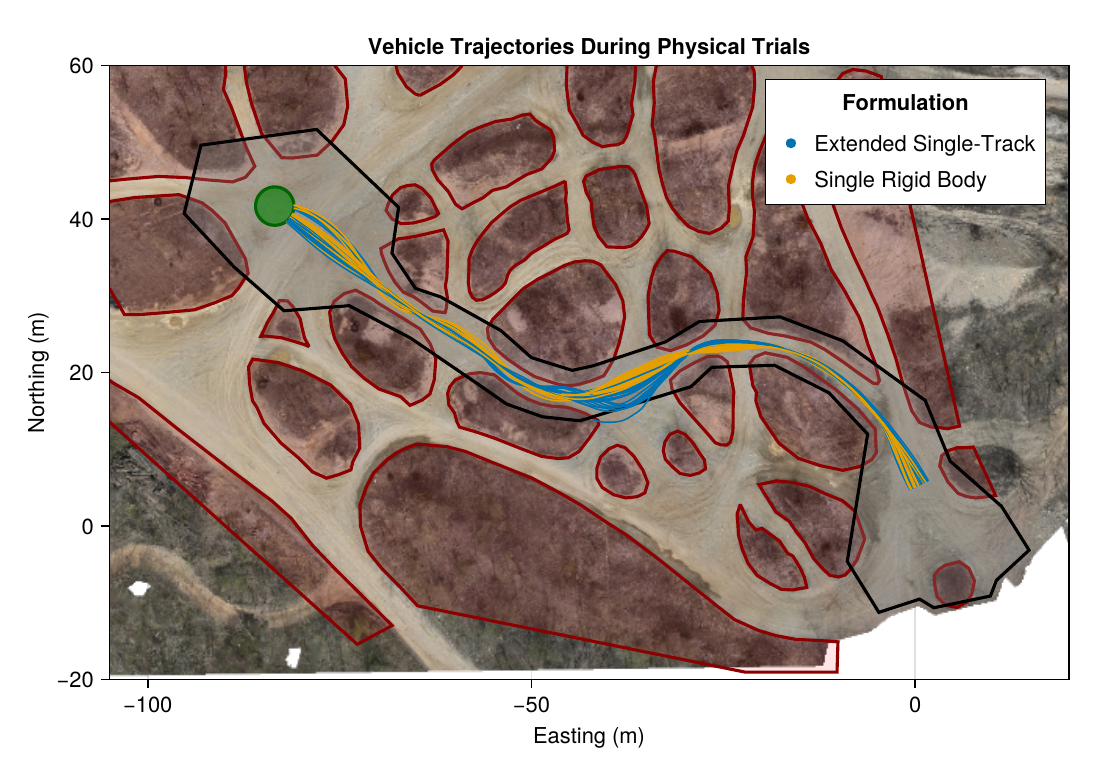}
    \caption{Measured vehicle trajectories during physical trials.
    The extended single-track formulation experienced one failure, where the vehicle exited the path boundary and collided with an obstacle.
    The single rigid body formulation's route is also more consistent, indicating greater authority over the vehicle's trajectory.}
    \label{fig:closed-loop:real:trajs}
\end{figure}

Out of 24 trials each, the extended single-track formulation experienced one failure while the single rigid body experienced none.
The outcomes are shown in Table \ref{tab:closed-loop:real:results}, and a composite time-lapse image of the failure is shown in Fig. \ref{fig:closed-loop:real:failure-composite}.
In the failed trial, the vehicle slips left after navigating the ridge, leading to inability to make the banked right turn required to stay on the primary path.
Instead, the vehicle begins to drive into the trees separating the primary and secondary paths, requiring a manual takeover and application of brakes to avoid collision.
As is visible in Fig. \ref{fig:closed-loop:real:trajs}, the trees are marked as obstacles, and as discussed in Sec. \ref{sec:formulation:ocp}, both formulations are equally charged with avoiding them.

In the failed trial, the vehicle experiences non-negligible vertical dynamics while navigating the ridge that occurs immediately after the hill.
Fig. \ref{fig:liftoff} is taken shortly before the failure and shows the tire liftoff that occurs along with rapid changes in pitch due to the terrain roughness.
As previously discussed, the extended single-track vehicle model does not model these dynamics: the vertical and suspension dynamics act as disturbances.
The inability to model and sufficiently compensate for these dynamics is believed to be the primary cause of failure.

\begin{table}
    \centering
    \caption{Outcomes of trials during physical experiments}
  \begin{tabular}{ccc}
        \hline
        Target Speed & Extended Single-Track & Single Rigid Body\\ \hline
        \qty{4.0}{\m\per\s} & 5 successes, 0 failures & 5 successes, 0 failures \\
        \qty{4.5}{\m\per\s} & 9 successes, 0 failures & 9 successes, 0 failures \\
        \qty{5.0}{\m\per\s} & 4 successes, 1 failure\phantom{s}  & 5 successes, 0 failures \\
        \qty{5.5}{\m\per\s} & 5 successes, 0 failures & 5 successes, 0 failures \\
    \end{tabular}
    \label{tab:closed-loop:real:results}
\end{table}

The measured vehicle trajectories for all physical trials are shown in Fig. \ref{fig:closed-loop:real:trajs}.
Notably, the extended single track formulation experiences consistent overshoot after the rough ridge and before the banked right turn.
Even though only one manual takeover occurs, when the CoM clearly violates the path boundary and collides with an obstacle, the plot demonstrates a consistent pattern:
    the extended single-track formulation's route is less consistent, indicating poorer authority over the vehicle's trajectory.
The single rigid body formulation is comparatively more consistent, and it does a superior job of keeping the vehicle aligned on the most direct trajectory to the goal.

\begin{figure}
    \centering
    \includegraphics[width=\linewidth]{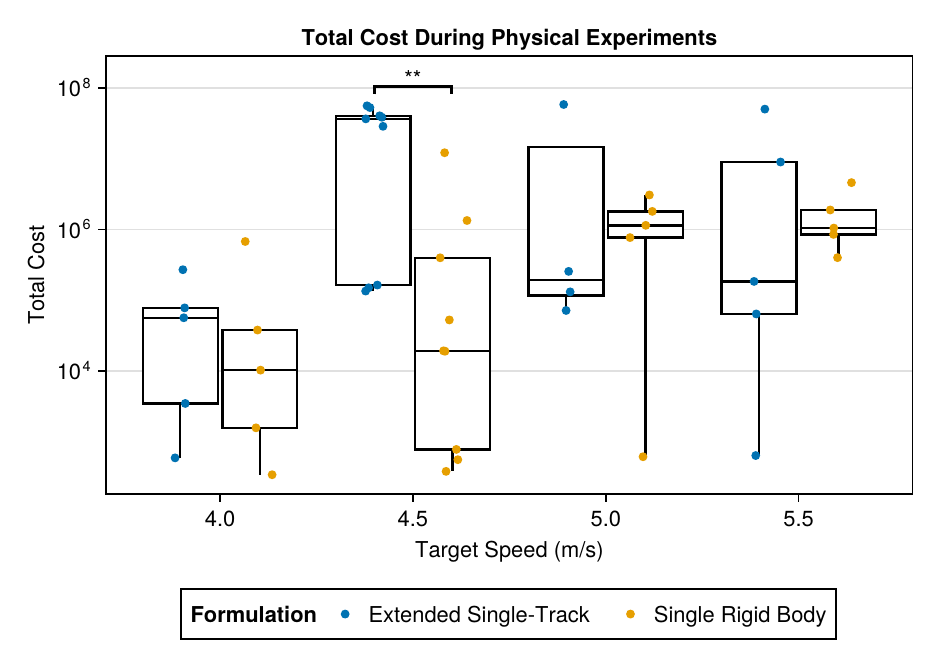}
    \caption{
        Total accrued costs during physical trials
        (note: the x-axis is categorical; deviation from the axis ticks is for visual clarity only).
        The single rigid body formulation has a statistically significant reduction in cost in the \qty{4.5}{\m\per\s} trials.}
    \label{fig:closed-loop:real:cost}
\end{figure}

Additionally, the total accrued cost of each successful trial is plotted in Fig. \ref{fig:closed-loop:real:cost}.
The cost formulation given in \eqref{eq:framework:cost} is integrated over a horizon equal to the full run-time of the trial.
Among a set of successful trials, this enables comparison of how well the vehicle completed the navigation task.

The primary contributor to the plotted costs is integrated obstacle violation.
The MPC formulation adds cost for each wheel that intersects with a set of inflated obstacles.
The amount of cost rises as more wheels are in contact, as the contact increases in duration, or as the depth of contact grows.
Fig \ref{fig:closed-loop:real:trajs} plots the trajectory of the CoM,
    so it delivers a less accurate view of obstacle violation compared to Fig. \ref{fig:closed-loop:real:cost}.

An exact Mann-Whitney U test is performed for each target speed, testing the null hypothesis that the cost measurements for each formulation come from the same distribution.
The null hypothesis is rejected at the $\alpha$ = \num{0.05} level for the \qty{4.5}{\m\per\s} trials.
This indicates the visible reduction in median cost delivered by the single rigid body formulation is statistically significant.
The p-value is \num{0.0040}.
No statistically significant differences are found at other speeds, likely due to the
decreased number of trials conducted at speeds other than \qty{4.5}{\m\per\s}.

\section{Conclusion}\label{sec:conclusion}
This work proposes an MPC formulation that uses an analytical vehicle dynamics model with increased degrees-of-freedom compared to similar approaches and an energy-based rollover-prevention constraint.
It aims to enable safe and capable local trajectory planning for high-speed off-road autonomous vehicles in rugged environments.
The formulation is made real-time feasible through GPGPU programming, and it is found to have a decreased proportion of rollovers and an increased proportion of successes compared to a \sota baseline, indicating it increases safety without reducing mobility.
Trials evaluate autonomous navigation performance at the extreme limits of mobility in a real off-road environment and comprise both simulated runs and full-scale physical experiments.

Additionally, \sota approaches leveraging planar vehicle models are analytically shown to only protect against rollovers caused by excessive handling on terrain with sufficient traction.
This type of rollover is inapplicable to some vehicle-terrain combinations,
    meaning imposing a \sota rollover-prevention constraint in these cases amounts to an unnecessary computational burden since it is numerically impossible to violate.
This is found to be true of the vehicle and terrain used in the experimental trials performed in this work.
If closed-loop rollover-prevention constraints are not required, e.g., due to low speed operation or mild terrain, the constraint may be dropped from the formulation.
Otherwise, the proposed approach may be used to provide such a constraint.

Several limitations of the proposed approach and analysis are identified, which may be addressed by future work.
Regarding the approach:
    First, the current formulation does not consider uncertainty.
        Several influential parameters, such as the terrain's coefficient of friction, may not be fully known and can only be estimated with a degree of uncertainty.
        In this setting, rollover occurrence is no longer a fixed outcome of control; instead it is an uncertain event with a certain risk.
        Adopting a stochastic MPC approach may yield superior ability to provide safe, extreme local trajectory plans.
    Second, the empirical-argmin optimizer does not leverage warm starting or weighted averaging.
        While this is demonstrated to be real-time feasible, methods using these techniques
            may be capable of providing similar or better performance with a smaller number of samples.
        This may enable faster update rates or implementation on resource-constrained hardware.

Regarding the analysis:
    First, longitudinal velocity is held constant, and the formulations attempt to minimize cost via steering only.
        At especially high speeds, it is possible that no steering policy is capable of preventing rollover, and an optimal policy would instead choose to decrease speed.
        Providing the formulations this option via simultaneous optimization of speed and steering would present additional context to evaluate their abilities to predict and mitigate risk.
    Second, the simulated trials provided the vehicle with perfect \textit{a priori} map knowledge and odometry measurements, and the physical experiments provided low-noise estimates of the same.
        However, this may not be the case for all fielded conditions:
            terrain and obstacles may need to be estimated online, and odometry estimators may not have access to RTK-corrected GNSS signals.
        Continued work on developing the simulation tools and experimental equipment may
            provide additional insight into the performance of these algorithms under more information-constrained fielded conditions.

\section{Acknowledgments}
The authors thank Junsik Eom and Austin Buchan for assistance in performing physical experiments, MTRI Inc. for photogrammetry services, and Martin Hirschkorn of Traxara Robotics for vehicle modeling support.
The authors acknowledge use of the Makie.jl plotting package \cite{danischMakiejlFlexibleHighperformance2021}.

\appendices
\section{Notation}\label{app:notation}
Vectors, denoted in bold, are resolved in one of two coordinate frames.
$\mathcal{W}$ is the world-fixed coordinate frame, with basis vectors $\mathbf{w}_x$, $\mathbf{w}_y$, and $\mathbf{w}_z$, where $\mathbf{w}_z$ points up.
$\mathcal{B}$ is the body-fixed coordinate frame, rigidly attached to the vehicle chassis.
It is spanned by $\mathbf{b}_x$, $\mathbf{b}_y$, and $\mathbf{b}_z$, which point forward, left, and up, relative to the chassis, respectively.

Left subscripts indicate the frame a vector is resolved in, and right subscripts indicate the component of such a vector:
\begin{align}
    \resveccomp{q}{F}{i} &\triangleq \bm{q} \cdot \mathbf{f}_i \\
    \resvec{q}{F} &\triangleq \begin{bmatrix}
        \resveccomp{q}{F}{x} & \resveccomp{q}{F}{y} & \resveccomp{q}{F}{z}
    \end{bmatrix}^T
\end{align}
where $\bm{q}$ is an arbitrary vector and $\mathcal{F}$ is an arbitrary frame.

Additionally, notation for common inertial and geometric parameters is summarized in Table \ref{tab:notation:params}.
\begin{table}
    \centering
    \caption{Notation for common inertial and geometric parameters}
    \begin{tabular}{cl}
        \hline Parameter & Meaning \\ \hline
        $M$ & vehicle mass \\
        $J_{xx}$ & vehicle moment of inertia about $\mathbf{b}_x$ \\
        $J_{yy}$ & vehicle moment of inertia about $\mathbf{b}_y$ \\
        $J_{zz}$ & vehicle moment of inertia about $\mathbf{b}_z$ \\
        $L_f$ & long. distance from CoM to front axle \\
        $L_r$ & long. distance from CoM to rear axle \\
        $e$ & wheel track \\
        $h$ & vertical distance from CoM to axles \\
        $R$ & tire radius \\
    \end{tabular}
    \label{tab:notation:params}
\end{table}

\section{Vehicle Model Parameters}\label{app:params}
Parameters describing the Polaris MRZR D4 ultra-light off-road vehicle used in this work are given in Table \ref{tab:params}.
For parameters common to both models, the same numerical values are used in each.
Geometric parameters are measured at rest on level ground, and inertial properties describe the entire vehicle.
Suspension parameters are found through a system-identification procedure.

\begin{table}[t]
    \centering
    \caption{Model parameters}
    \begin{tabular}{cl}
        \hline Parameter & Value \\ \hline
        $M$ & \qty{969}{\kg} \\
        $J_{xx}$ & \qty{280.9}{\kg\m\squared} \\
        $J_{yy}$ & \qty{692.1}{\kg\m\squared} \\
        $J_{zz}$ & \qty{810.7}{\kg\m\squared} \\
        $L_f$ & \qty{1.565}{\m} \\
        $L_r$ & \qty{1.148}{\m} \\
        $e$ & \qty{1.280}{\m} \\
        $h$ & \qty{0.380}{\m} \\
        $R$ & \qty{0.291}{\m} \\
        $k_f$ & \qty{4.2e4}{\N\per\m} \\
        $k_r$ & \qty{5.8e4}{\N\per\m} \\
        $b_f$ & \qty{3.1e3}{\N\s\per\m} \\
        $b_r$ & \qty{4.3e3}{\N\s\per\m} \\
        $\delta_{\mathrm{max}}$ & \qty{0.639}{\radian} \\
        $\dot{\delta}_{\mathrm{max}}$ & \qty{1}{\radian\per\s}
    \end{tabular}
    \label{tab:params}
\end{table}

In addition, each model requires parameters for the tire model.
The simulated experiments use the parameters given in Table \ref{tab:params:sim},
    which are found through a model fitting process performed on the fine-grained dry soil tire model used within the plant simulator.
The plant tire model is fit to data from physical experiments performed at Keweenaw Research Center in Calumet, MI.
The physical experiments use the parameters given in Table \ref{tab:params:real},
    which are found through a model fitting process performed on the data-driven tire model used in \cite{yuRealTimeTerrainAdaptiveLocal2025}.
The data-driven tire model's training data consists of measurements from physical experiments performed with a highly-similar vehicle in the same off-road testing location.

\begin{table}[t]
    \centering
    \caption{Tire parameters for simulated trials}
    \begin{tabular}{cl}
        \hline Parameter & Value \\ \hline
        $C$ & \qty[per-mode=power]{6.1}{\per\radian} \\
        $\mu$ & 0.6
    \end{tabular}
    \label{tab:params:sim}
\end{table}

\begin{table}[t]
    \centering
    \caption{Tire parameters for physical trials}
    \begin{tabular}{cl}
        \hline Parameter & Value \\ \hline
        $C$ & \qty[per-mode=power]{1.7}{\per\radian} \\
        $\mu$ & 0.4
    \end{tabular}
    \label{tab:params:real}
\end{table}

\section{Open-Loop Analysis}\label{app:open-loop}

\begin{figure}[ht]
    \centering
    \includegraphics[width=\linewidth]{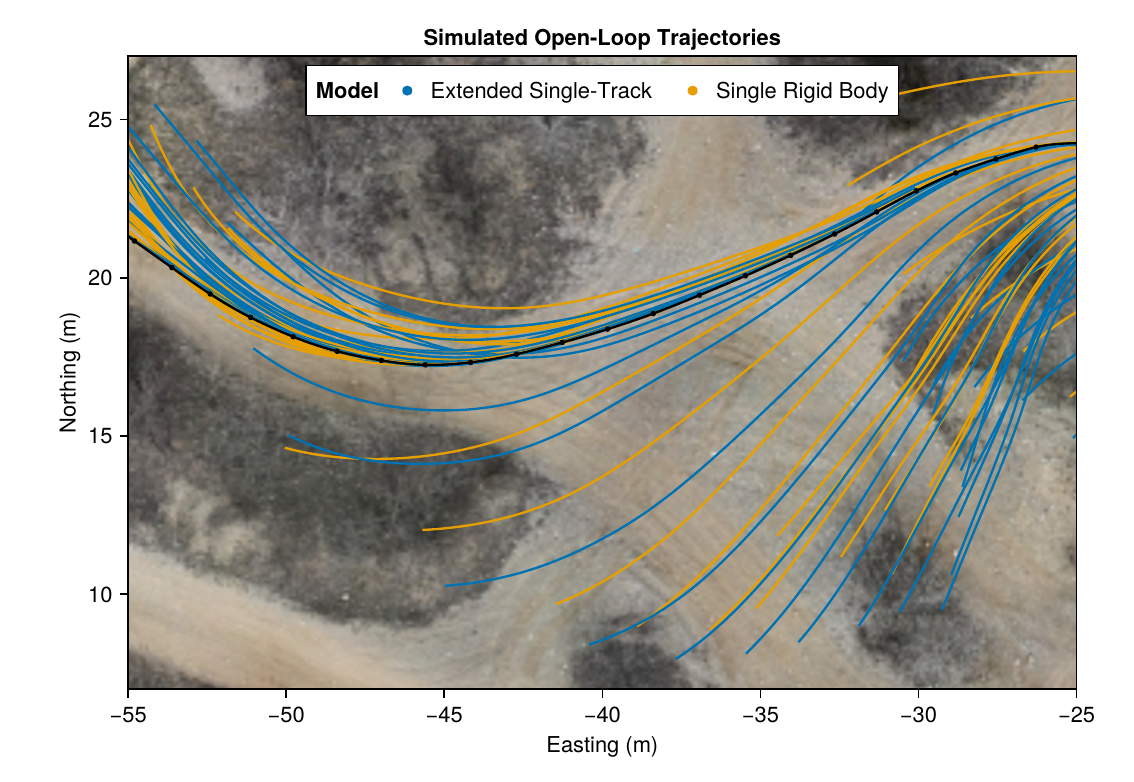}
    \caption{
        Subset of simulated trajectories used for open-loop analysis,
        focusing on the ridge and turn included in scenario 1
        where the extended single-track formulation
        experienced a failure during physical testing.
        The plant trajectory is shown as a black line, open-loop
        model predictions are shown as colored lines,
        and black points mark the locations where the former originate.
        The trajectories display decreased overshoot error by the single rigid body model.}
    \label{fig:open-loop:traj}
\end{figure}

\num{52146} total simulated trajectories are used to evaluate the open-loop performance of the two models studied in this work.
Each trajectory is created by simulating both models and the high-fidelity plant with identical initial conditions and controls.
Each simulation lasts \qty{4}{\s}, the duration of the prediction horizon,
and the initial conditions and controls leverage the plant trajectories generated by the simulated closed-loop analysis described in Sec. \ref{sec:results:sim}.
Specifically, scenario 1 trajectories conducted at \qty{7.5}{\m\per\s} and terminating in success are used.
A subset of the open-loop trajectories are plotted in Fig. \ref{fig:open-loop:traj}.

Error is computed for location and ESM, respectively, via
\begin{align}
    \text{location error} &= \frac{1}{t_f-t_0}\int_{t_0}^{t_f} \sqrt{(\hat{x}-x)^2+(\hat{y}-y)^2} dt \\
    \text{ESM error} &= \frac{1}{t_f-t_0}\int_{t_0}^{t_f} \lvert \hat{U}_\mathrm{ESM} - U_\mathrm{ESM} \rvert dt
\end{align}
where $\hat{x}$ and $x$ refer to values computed by the model and plant, respectively, and other variables are similarly denoted.

The testing found 10.1\% and 3.1\% relative improvements in median prediction error for location and ESM, respectively,
when using the single rigid body model over the extended single-track model.
An approximate Mann-Whitney U test finds that both error reductions are statistically significant, with p-values less than 0.0001 in each case.
However, it is emphasized that the primary advantage of the single rigid body formulation
over the extended single-track formulation is the increased ability to guarantee safety against rollover.
Open-loop performance improvements, especially to non-rollover-related states such as location, are supplementary to this.

\bibliographystyle{IEEEtran}
\bibliography{references}

\vskip -2\baselineskip plus -1fil
    \begin{IEEEbiography}[{\includegraphics[width=1in,height=1.25in,clip,keepaspectratio]{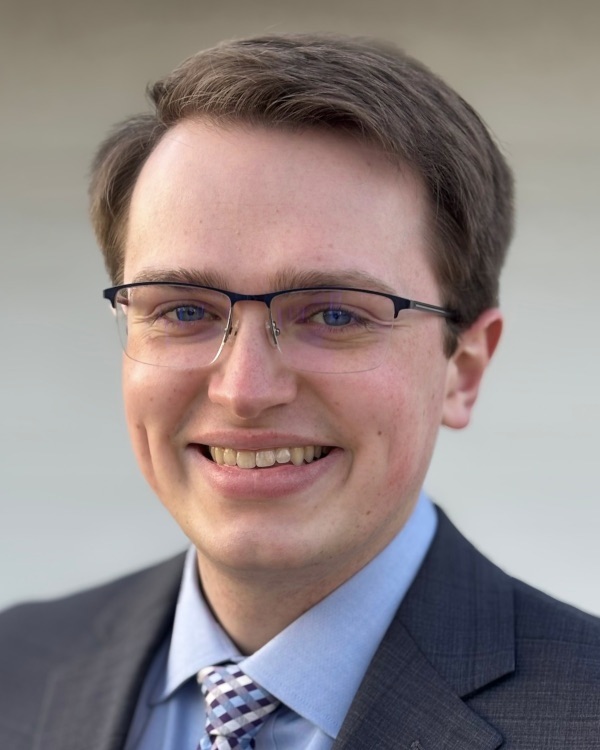}}]{James R. Baxter}
    received the B.Sc. degree in mechanical engineering from Purdue University, West Lafayette, IN, USA, in 2022,
    and the M.S.E. degree in mechanical engineering from the University of Michigan, Ann Arbor, MI, USA, in 2024, where he is currently working toward the Ph.D. degree in mechanical engineering.
    His research interests include motion planning and control for highly dynamical systems, with applications to autonomous robotics.
    \end{IEEEbiography}
\vskip -2\baselineskip plus -1fil
    \begin{IEEEbiography}[{\includegraphics[width=1in,height=1.25in,clip,keepaspectratio]{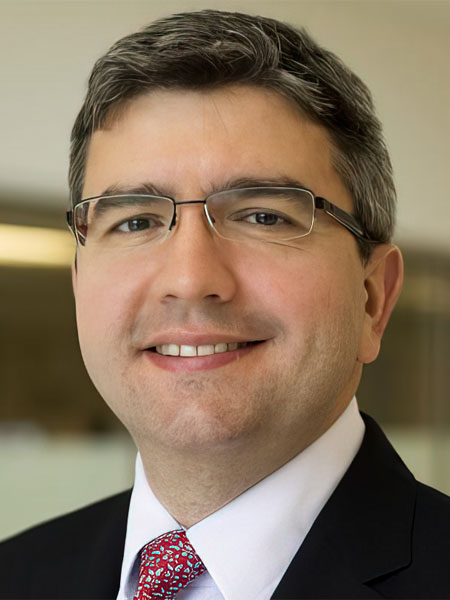}}]{Bogdan I. Epureanu}
    received the Ph.D. degree in mechanical engineering from Duke University, Durham, NC, USA, in 1999.
    He is currently the Roger L. McCarthy Professor and an Arthur F. Thurnau Professor with the Department of Mechanical Engineering, University of Michigan, Ann Arbor, MI, USA.
    He has a courtesy appointment in electrical engineering and computer science, and he is the Director of the Automotive Research Center.
    His research focuses on nonlinear dynamics of complex systems, such as teaming of autonomous vehicles, enhanced aircraft safety and performance, and aeromechanics.
    \end{IEEEbiography}
\vskip -2\baselineskip plus -1fil
    \begin{IEEEbiography}[{\includegraphics[width=1in,height=1.25in,clip,keepaspectratio]{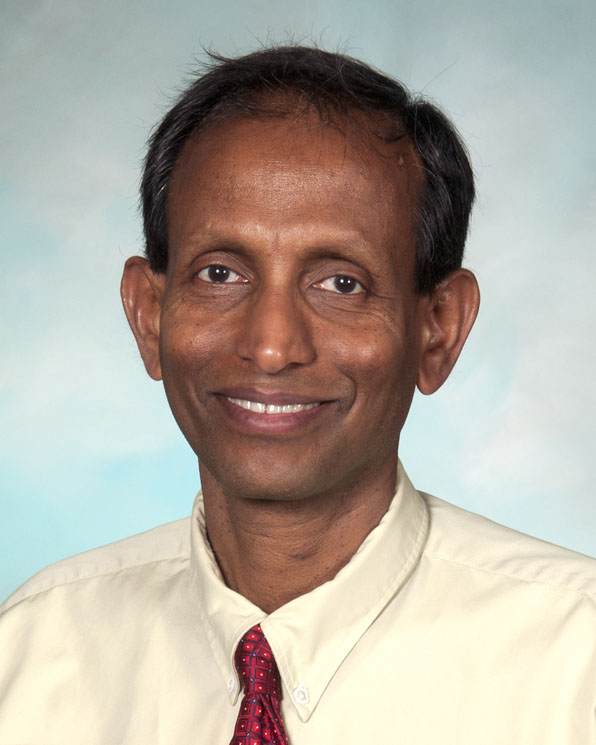}}]{Paramsothy Jayakumar}
    received the M.S. and Ph.D. degrees in structural dynamics from California Institute of Technology, Pasadena, CA, USA,
    and the B.S.E. degree from the University of Peradeniya, Sri Lanka.
    He is a Senior Technical Expert in Analytics at U.S. Army Ground Vehicle Systems Center. He is a Fellow of SAE and ASME.
    He is an Associate Editor of the ASME Journal of Autonomous Vehicles and Systems, and Editorial Board Member of the International Journal of Vehicle Performance and the ISTVS Journal of Terramechanics.
    \end{IEEEbiography}
\vskip -2\baselineskip plus -1fil
    \begin{IEEEbiography}[{\includegraphics[width=1in,height=1.25in,clip,keepaspectratio]{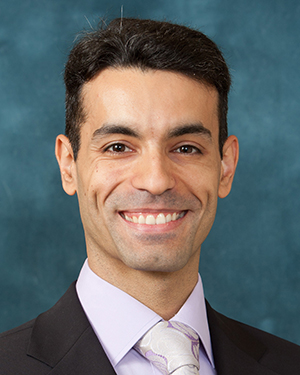}}]{Tulga Ersal}
    received the B.S.E. degree from the Istanbul Technical University, Istanbul, Turkey, in 2001, and the M.S. and Ph.D. degrees from the University of Michigan, Ann Arbor, MI, USA, in 2003 and 2007, respectively, all in mechanical engineering.
    He is currently an Associate Professor with the Department of Mechanical Engineering, University of Michigan. His research interests include modeling, simulation, and control of dynamic systems, with applications to vehicle and energy systems.
    \end{IEEEbiography}

\end{document}